\documentclass[11pt]{article}

\usepackage[final]{acl}

\usepackage{times}
\usepackage{latexsym}
\usepackage[T1]{fontenc}
\usepackage[utf8]{inputenc}
\usepackage{microtype}
\usepackage{inconsolata}
\usepackage{graphicx}

\usepackage{enumitem}
\usepackage{amsmath}
\usepackage{amssymb}

\usepackage{comment}
\usepackage{booktabs}
\usepackage{subcaption}
\usepackage{xspace}
\usepackage{multirow}
\usepackage{makecell}
\usepackage{float}
\usepackage{stfloats}
\usepackage{algorithm}
\usepackage{algpseudocode}

\newif\ifshowcomments
\showcommentsfalse

\long\def\comment#1{}
\long\def\comments#1{}

\newcommand{\ours}{CoinRAG\xspace}

\newcommand{\standardRAG}{Standard RAG\xspace}
\newcommand{\standardCAG}{Standard CAG\xspace}

\newcommand{\baseLLM}{Qwen2-7B-Instruct\xspace}
\newcommand{\embed}{BGE-M3\xspace}

\title{\ours: Contextualized Information Nugget KV Cache Reuse \\for Long-Context RAG}

\author{
Gyuwan Kim$^{1}$ \quad Cheoneum Park$^{2}$ \quad Tao Yang$^{1}$ \\
$^{1}$University of California, Santa Barbara \quad $^{2}$Hanbat National University\\
\texttt{\{gyuwankim, tyang\}@ucsb.edu, parkce@hanbat.ac.kr}
}

\begin{document}

\maketitle

\begin{abstract}
Recent optimization studies on Retrieval-Augmented Generation (RAG) have exploited chunk-level KV cache reuse to avoid processing long retrieved contexts for higher efficiency, while significant information redundancy and noise still remain in the coarse-grained chunks.
This paper optimizes the Pareto frontier under low prefill latency constraints while maximizing accuracy by proposing \ours{} (\textbf{Co}ntextualized \textbf{I}nformation \textbf{N}ugget KV Cache Reuse for Long-Context \textbf{RAG}).
The name metaphorically reflects our core mechanism: much like assembling small tokens (or ``coins'') to accumulate a larger value, \ours{} compositionally reuses offline-computed, fine-grained nugget caches to form a learned contextual representation efficiently in a more semantically relevant but compact manner.
Specifically, instead of full-chunk encoding, \ours{} identifies query-relevant semantic units within retrieved chunks through two-stage retrieval and seamlessly assembles their 
sliced KV representations with a chunk-level context.
\ours{} can be integrated with CacheBlend's selective recomputation for further accuracy gains under the same latency budget.
Extensive evaluations on LongBench multi-hop question answering tasks demonstrate that \ours{} substantially shortens the active context and outperforms the other baselines with a new Pareto frontier and an average 5.3\% relative improvement in answer quality (F1) under a standard fast prefill latency budget.

\end{abstract}

\setlength{\abovedisplayskip}{4pt}
\setlength{\belowdisplayskip}{4pt}
\setlength{\abovedisplayshortskip}{2pt}
\setlength{\belowdisplayshortskip}{2pt}

\section{Introduction}

Retrieval-Augmented Generation (RAG)~\citep{lewis2020retrieval, gao2023retrieval} is a widely adopted paradigm for enhancing large language models (LLMs) with external knowledge~\citep{chen2024benchmarking}.
By feeding retrieved documents to the LLM with the query, RAG improves factual consistency~\citep{gao2023enabling} and reduces hallucination~\citep{shuster2021retrieval} without model retraining.
However, when RAG systems operate at the coarse granularity of full documents or large text chunks, inference is expensive, leading to high prefill latency and computational redundancy.
To improve inference efficiency, a promising remedy is to precompute and reuse key-value (KV) caches~\citep{pope2023efficiently}, a technique that has been extensively engineered for prompt management and multi-query prefix sharing~\citep{kwon2023efficient, zheng2024sglang, gim2024prompt, haoyang2025survey}.
Crucially, this core mechanical capability underpins the emerging \textit{Cache-augmented Generation} (CAG) paradigm~\citep{chan2025don}, which proposes pre-loading entire external knowledge bases into the model's context to eliminate real-time retrieval.
RoPE-based position rotation~\citep{su2024roformer, lu2025turborag, yang2025kvlink} has been developed further to capture relative positional contexts of dynamically retrieved chunks, enabling chunk-level KV cache reuse during inference.

As RAG has become popular for user-facing web applications, inference must answer questions at interactive latencies while sustaining the high-throughput demands of large live query traffic.
Commercial web services operate under a Service Level Agreement (SLA) that specifies tail latency (percentiles) rather than average latency.
Response time for such services is the time to deliver the first byte of a response, and 100 ms or less is perceived as instantaneous~\citep{PubNub-responsetime, 1968ACM-Miller-ResponseTime}.
Prior work on low-latency machine learning inference has likewise pursued a P99 tail latency budget within 100 ms~\citep{2015SIGIR-tailLatency, 2017NSDI-Clipper-Online100ms, 2020OSDI-low-latency100ms}.
Our study therefore adopts a Time-to-First-Token (TTFT) budget of P99 latency under 100 ms.
Note that lowering latency also leads to a higher throughput.

Standard RAG, even with KV cache reuse, is expensive in terms of TTFT. 
To meet SLA under a low-latency budget, it needs to limit the retrieval scope, degrading answer accuracy.
This paper focuses on maximizing RAG accuracy under a low-latency budget and takes a new approach by operating RAG at a finer-grained level of information extracted in advance from text chunks with a slim representation, which also reduces the risk that relevant evidence gets lost amid long, noisy contexts~\citep{liu2024lost}.
Specifically, we extract information nuggets from text chunks offline, instead of retrieving and encoding them entirely at inference time.
Our approach is motivated by prior information retrieval studies, which use the human-identified {\em nuggets} for answer evaluation~\citep{2003nugget} and which extract nuggets dynamically from each document via an LLM~\citep{pradeep2025great, lajewska2025ginger}.

This paper makes the following contributions.
1) It proposes a RAG framework for efficient inference with a slim nugget-based representation for contextual KV-cache reuse, targeted for higher accuracy under a standard fast response-time budget.
2) This scheme comes with the following techniques: (a) Offline extraction of text-span-based nuggets, nugget-aware fine-tuning, and 
two-stage online retrieval that selects relevant nuggets from many candidates; (b) Contextualized KV cache composition 
for query-relevant nuggets from the same chunks.
(c) An integration of \ours{} with CacheBlend~\citep{yao2025cacheblend} that further improves accuracy through selective cross-chunk interaction.

We evaluate \ours{} on multi-hop question answering benchmarks from LongBench~\citep{bai2024longbench}, where it consistently outperforms \standardRAG{} and \standardCAG{} with a 5.3\% (41.7 vs. 39.6) answer quality improvement averaged over 3 datasets under the P99 latency 100 ms budget.
When removing the latency limit entirely, \ours{} still outperforms 2 out of 3 datasets with a 3-dataset average F1 score improvement of 5.2\%.

\section{Proposed Method: \ours{}}
\label{sec:proposed_method}

We propose \ours{} (\textbf{Co}ntextualized \textbf{I}nformation \textbf{N}ugget KV Cache Reuse for Long-Context \textbf{RAG}), motivated by two key design considerations for higher RAG accuracy under a low-latency budget.
First, the granularity of KV cache at the chunk level is still too coarse, often adding noise and unnecessary redundancy.
We instead seek a slim representation of essential content in text chunks that maximizes the information density delivered to the LLM, representing each document chunk with a set of text spans capturing essential information nuggets.
Second, we take a query-driven nugget retrieval approach as dynamic filtration to select relevant nugget slices with contextual KV cache composition, preventing attention dilution with enhanced reasoning precision.

Appendix~\ref{app:inference} provides a step-by-step inference walkthrough example for \ours{}.
Appendix~\ref{app:inference-cases} shows examples of answer generation that succeeded and failed with \ours{}.

\subsection{Problem Formulation}
\label{sec:problem_formulation}

In standard RAG, a large corpus $\mathcal{D}$ consists of long documents pre-segmented into fixed-size text chunks $\mathcal{B} = \{b_j\}$, where each chunk is a sequence of tokens.
Traditional frameworks generate an answer to a query $q$ by jointly encoding the system prompt $p$, retrieved chunks, and the query $q$ online during the prefill stage.
This design introduces significant computational redundancy and high prefill latency because long contexts are repeatedly re-encoded across queries.

To eliminate this bottleneck, \ours{} shifts context encoding entirely offline.
During this phase, every chunk $b_j \in \mathcal{B}$ is processed through a one-time forward pass to compute and store its full-context key-value (KV) representations $C_{b_j}$.
At online inference, \ours{} bypasses full-text re-encoding.
Instead of computing hidden states from raw text, our framework dynamically extracts query-relevant textual units within retrieved chunks, slices their corresponding precomputed KV caches, and aligns them into a single prefix context cache $C_{\mathrm{ctx}}$.
The online query is then encoded conditioned on this composed prefix cache, and the final context cache is constructed by concatenating $C_{\mathrm{ctx}}$ with the newly computed query representations: $\mathrm{KV}_{\mathrm{\ours{}}} = C_{\mathrm{ctx}} \oplus \mathrm{KV}_{\mathcal{M}}(q; C_{\mathrm{ctx}})$, where $\oplus$ denotes KV cache concatenation along the sequence dimension, and $\mathrm{KV}_{\mathcal{M}}(q; C_{\mathrm{ctx}})$ represents the query encoding pass.

Appendix~\ref{app:generation_prompt} gives the prompt template injected by \ours{} into an LLM for answer generation with the retrieved nuggets and the input query.

\subsection{Offline Nugget Extraction}
\label{sec:nugget_offline}

Unlike prior nugget-based approaches that generate nuggets online via an LLM and encode isolated nuggets without context, \ours{} extracts candidate nuggets from each text chunk $b_j \in \mathcal{B}$ offline, identifying each as a contiguous span defined by start and end token indices within its source chunk.
This span-level grounding lets us later slice the corresponding portion of the precomputed chunk cache to form a contextualized nugget KV cache, rather than needing to re-encode the nugget's text in isolation.

\begin{algorithm}[htbp]
\caption{Nugget extraction pipeline.}
\label{alg:nugget}
\footnotesize
\begin{algorithmic}[1]
\Require passage $p$; LLM extractor $\mathcal{E}_{\text{LLM}}$; similarity $\mathrm{sim}$; threshold $\tau$
\Ensure nugget set $\mathcal{N}(p)$, expressed as token span indices
\State $\{\tilde{n}_1, \ldots, \tilde{n}_K\} \leftarrow \mathcal{E}_{\text{LLM}}(p)$ \Comment{LLM prompting}
\State $\mathcal{N}(p) \leftarrow \emptyset$
\For{$i = 1, \ldots, K$}
  \If{$\tilde{n}_i \in \mathrm{Sub}(p)$} \Comment{Stage A: exact match}
    \State $n_i \leftarrow \tilde{n}_i$
  \Else \Comment{Stage B: fuzzy match}
    \State $W \leftarrow$ whole-word spans of $p$ near length $|\tilde{n}_i|$
    \State $s^{\ast} \leftarrow \arg\max_{w \in W} \mathrm{sim}(w, \tilde{n}_i)$
    \If{$\mathrm{sim}(s^{\ast}, \tilde{n}_i) \geq \tau$}
      \State $n_i \leftarrow s^{\ast}$
    \Else
      \State \textbf{continue} \Comment{Stage C: abandon}
    \EndIf
  \EndIf
  \State $(s_i, e_i) \leftarrow \texttt{find}(n_i, p)$ \Comment{token span}
  \State $\mathcal{N}(p) \leftarrow \mathcal{N}(p) \cup \{ (n_i, s_i, e_i) \}$
\EndFor
\State \Return $\mathcal{N}(p)$
\end{algorithmic}
\end{algorithm}

Algorithm~\ref{alg:nugget} delineates this extraction procedure applied to a raw text passage $p$.
An LLM first proposes candidate nuggets from $p$ (Line 1).
Each candidate is matched back to a token span in $p$: we first check whether it appears verbatim as a substring of $p$ (Stage A: exact match; Lines 4--5).
If not, we search word-level spans of $p$ near the candidate's length and match against the one with the highest similarity to the candidate, accepting it only if this similarity exceeds a threshold $\tau$ (Stage B: fuzzy match; Lines 6--10), and otherwise dropping the candidate (Stage C: abandon; Line 12).
These accepted spans, marked by their start and end token positions in $p$, form the extracted nuggets (Lines 15--16).
Appendix~\ref{app:nugget} provides the LLM prompt for extraction, an extraction example, and characteristics of this extraction on the test datasets.

\subsection{Two-Stage Online Nugget Retrieval}
\label{sec:nugget_online}

At runtime, our framework efficiently identifies the most informative segments for a given query through two-stage retrieval, scalable to large corpora.
Formally, we define a ranked list of the top-$k$ information nuggets for the query $q$ as $\mathcal{R}(q) = \{(b_i, s_i, e_i)\}_{i=1}^k$, where each tuple specifies the source chunk $b_i$ and the boundary token indices $[s_i, e_i]$ that define the contiguous span of the nugget within that chunk.
Given a query $q$, a dense retriever first fetches the top-$k_c$ text chunks from the corpus to construct a retrieved subset $\mathcal{B}_{\text{ret}} = \{b_i\}_{i=1}^{k_c} \subset \mathcal{B}$.
We then evaluate and rank the pre-extracted candidate nuggets nested exclusively within these retrieved chunks $\mathcal{B}_{\text{ret}}$ based on their embedding similarity to $q$, keeping the top-$k$ entries.

Crucially, rather than treating these extracted nuggets as isolated text snippets that lack semantic grounding, \ours{} utilizes the boundary indices $[s_i, e_i]$ as pointers to slice the corresponding portions of the precomputed offline chunk cache $C_{b_i}$.
This operation yields contextualized nugget KV caches, denoted as $C_{b_i}[s_i:e_i]$, ensuring that the representations inherently retain the deep, document-level context established during full-chunk preprocessing.
Because $C_{b_i}$ is computed by encoding the entire chunk $b_i$ before any nugget is selected, every token's KV state within the sliced range $[s_i, e_i]$ is identical to what it would be under a fresh encoding of the chunk, carrying the same contextual grounding that full-chunk encoding provides.
Isolated nugget encoding forfeits exactly this: re-encoding a nugget's raw text on its own recomputes its KV states from scratch, without access to the rest of the chunk.
\S\ref{sec:ablation} empirically confirms the benefit of this preserved conditioning by comparing contextualized nugget KV slicing against isolated nugget re-encoding.

\subsection{Contextualized KV Cache Composition with Position Alignment}
\label{sec:cache_composition}

Once loaded and sliced, the contextualized nugget caches are composed into a unified prefix cache $C_{\mathrm{ctx}}$ alongside the system prompt cache $C_p$.
Because individual nugget caches originate from different source chunks and token positions, their relative rotary position encodings (RoPE)~\citep{su2024roformer} mismatch, preventing naive concatenation.
To resolve these positional gaps, we adapt a position rotation operator $\mathrm{Rot}(C; \Delta)$ that shifts the positional indices of a cached KV block by an offset $\Delta$.

The unified context cache is constructed through granular cache composition:
$C_{\mathrm{ctx}} = C_p \oplus \mathrm{Rot}(C_{b_1}[s_1:e_1]; \Delta_1) \oplus \cdots \oplus \mathrm{Rot}(C_{b_k}[s_k:e_k]; \Delta_k)$.
To manage the structural alignment of these heterogeneous spans, \ours{} employs an \textit{order-preserving, contiguous position alignment strategy} that preserves the original document sequence order while packing nuggets consecutively to eliminate structural gaps.
Accordingly, the position rotation offset $\Delta_i$ for the $i$-th nugget is dynamically calculated as $\Delta_i = |p| + \sum_{j < i} (e_j - s_j + 1)$.
By masking non-retrieved intermediate text and assigning continuous positional indices, this mechanism minimizes the total context length, thereby reducing GPU memory footprints and accelerating decoding without altering the model architecture.

\subsection{Nugget-Aware Fine-tuning}
\label{sec:finetuning}

While \ours{} can operate in a training-free manner,
the online synthesis of non-contiguous cached segments inherently introduces a structural training-inference gap, as LLMs are trained on continuous sequences.
To mitigate this 
gap and optimize downstream answer generation, \ours{} 
incorporates 
a \textit{nugget-aware fine-tuning} stage that aligns the training structure with test-time operations.

We train the LLM on a question-answering dataset where each instance consists of a query $q$, a target answer $y$, and a document collection containing a mixture of ground-truth and distractors.
To replicate the online extraction and composition pipeline, we fetch the top-$k$ relevant nuggets for $q$ from these documents and dynamically construct the prefix context cache $C_{\mathrm{ctx}}$ via the position alignment described in Section~\ref{sec:cache_composition}.
The optimization objective directly minimizes the standard cross-entropy loss on the target answer tokens using the model parameters: $\mathcal{L}(\mathcal{M}) = - \sum_{t=1}^T \log P_{\mathcal{M}}(y_t \mid y_{<t}, C_{\mathrm{ctx}}, q)$.

\subsection{Integration with CacheBlend}
\label{sec:cacheblend_integration}

Nuggets from different chunks do not attend to each other in $C_{\mathrm{ctx}}$, as each chunk is encoded independently.
\ours{} can be integrated with the selective recomputation of CacheBlend~\citep{yao2025cacheblend} to restore this cross-chunk interaction (Figure~\ref{fig:comparison}(f)).
After composing $C_{\mathrm{ctx}}$, we recompute the KV states of a fraction $r$ of its tokens with full attention over the preceding prefix, selecting the tokens whose KV states deviate most from full attention, following CacheBlend's criterion.
This requires no additional training, and $r$ trades prefill computation for accuracy.
We also tried integrating \ours{} with KVLink~\citep{yang2025kvlink}, which yielded mixed results (Appendix~\ref{app:cross_chunk}).
A potential reason is that KVLink's link tokens are designed to attend to all tokens in the preceding chunks, whereas in \ours{} they can attend only to the few short nugget spans retained from each chunk.

\section{Comparison to Prior RAG Paradigms}
\label{sec:comparison}

\begin{figure*}[t]
\centering
\begin{subfigure}{0.32\linewidth}
\centering
\includegraphics[width=\linewidth]{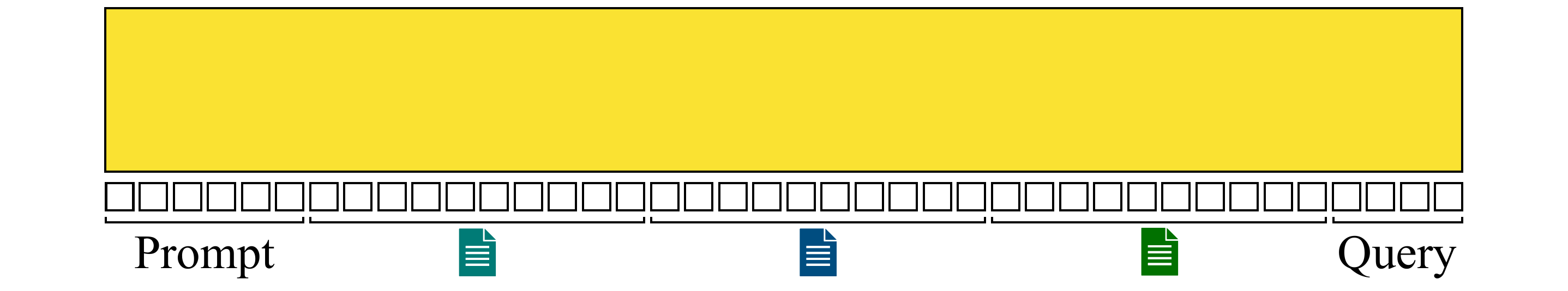}
\caption{\standardRAG{}}
\label{fig:standard_rag}
\end{subfigure}
\hfill
\begin{subfigure}{0.32\linewidth}
\centering
\includegraphics[width=\linewidth]{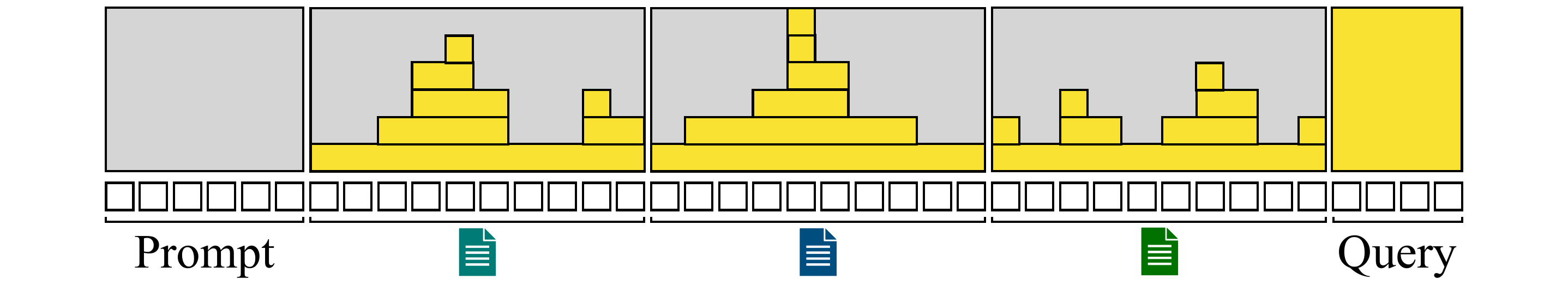}
\caption{CacheBlend}
\label{fig:cacheblend}
\end{subfigure}
\hfill
\begin{subfigure}{0.32\linewidth}
\centering
\includegraphics[width=\linewidth]{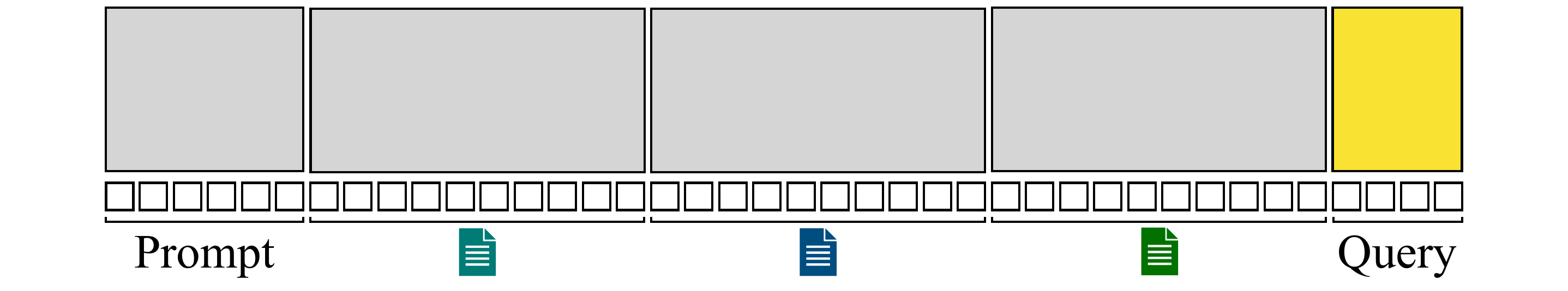}
\caption{TurboRAG}
\label{fig:turborag}
\end{subfigure}

\begin{subfigure}{0.32\linewidth}
\centering
\includegraphics[width=\linewidth]{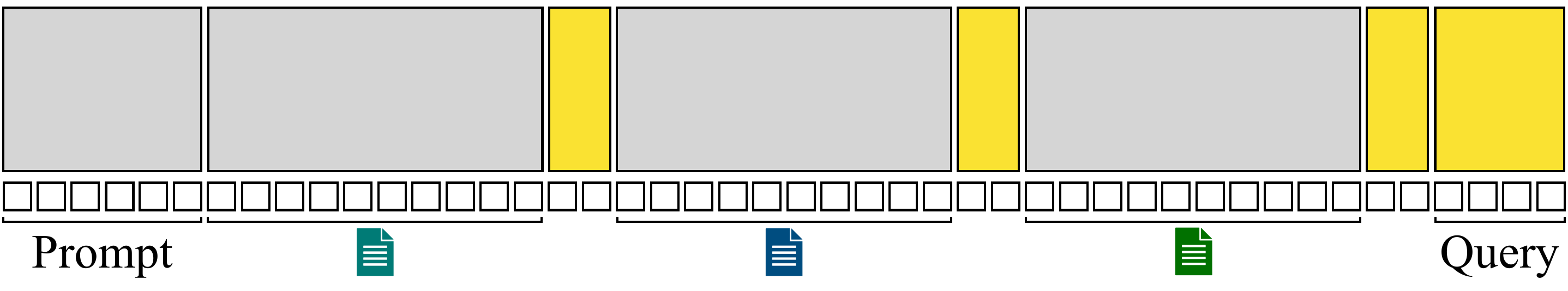}
\caption{KVLink}
\label{fig:kvlink}
\end{subfigure}
\hfill
\begin{subfigure}{0.32\linewidth}
\centering
\includegraphics[width=\linewidth]{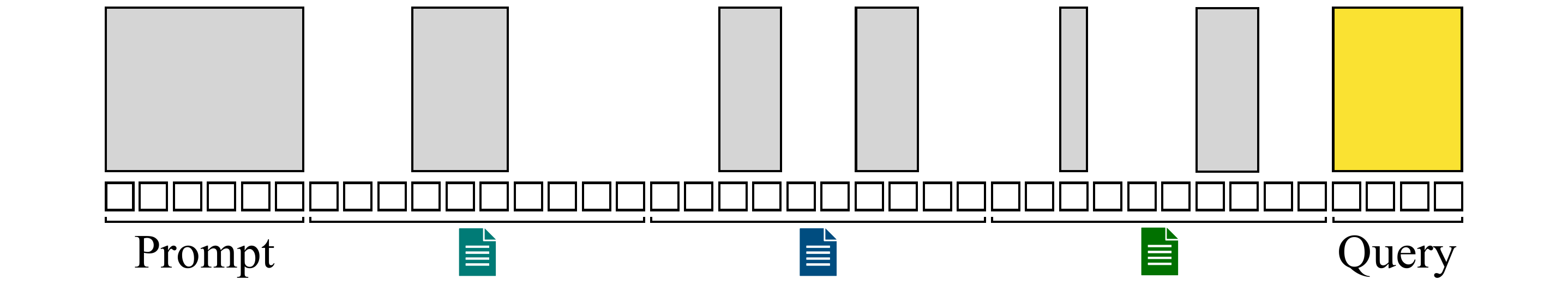}
\caption{\ours{}}
\label{fig:ours}
\end{subfigure}
\hfill
\begin{subfigure}{0.32\linewidth}
\centering
\includegraphics[width=\linewidth]{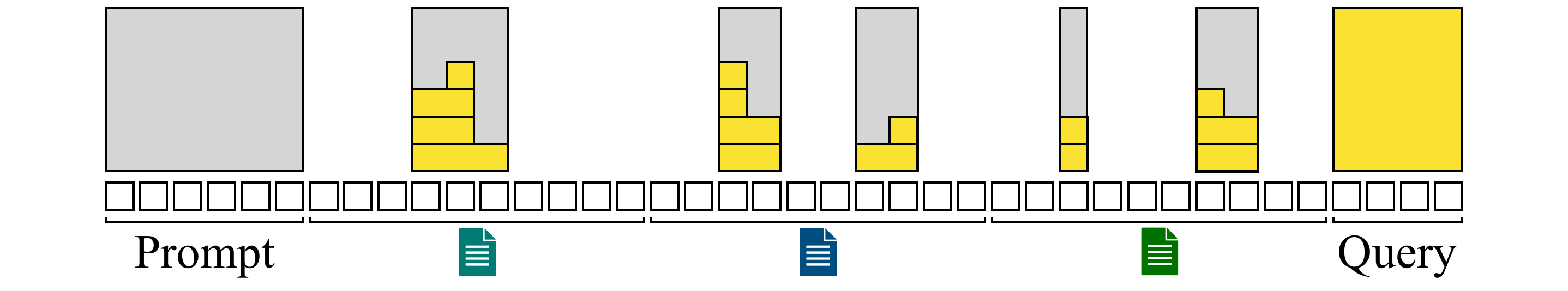}
\caption{\ours{}+CacheBlend}
\label{fig:ours_cacheblend}
\end{subfigure}

\vspace{0.3em}
\includegraphics[width=0.6\linewidth]{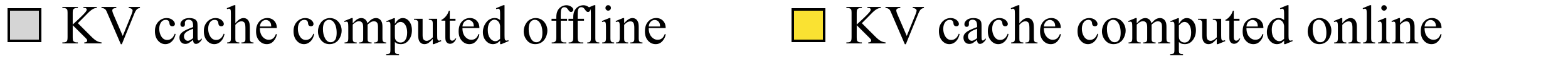}

\caption{
Comparison of context construction during the prefill stage across RAG paradigms.
Gray blocks represent KV caches precomputed \textit{offline}, where each token attends to all previous tokens within the same chunk, and yellow blocks represent KV caches computed \textit{online} during inference, where each token attends to all previous tokens.
}
\label{fig:comparison}
\end{figure*}

We position \ours{} against two lines of prior work: cache-augmented generation (\S\ref{sec:rel_cag}), which precomputes and reuses KV caches at the chunk level, and nugget-based RAG (\S\ref{sec:rel_nugget}), which represents retrieved evidence as fine-grained information nuggets.
Figure~\ref{fig:comparison} illustrates how each constructs the prefill KV cache, and Tables~\ref{tab:comparison} and~\ref{tab:nugget} summarize the comparison.

\textbf{\standardRAG{}} (Figure~\ref{fig:comparison}(a)) computes the interaction of full retrieved text and query entirely online via $\mathrm{KV}_{\text{StandardRAG}} = \mathrm{KV}_{\mathcal{M}}(p \oplus t_{\text{chunk}} \oplus q)$, where $t_{\text{chunk}} = b_1 \oplus \dots \oplus b_{k_c}$ concatenates the full chunks. It therefore suffers severe prefill latency from full online forward propagation over long sequences for every query.

\subsection{Cache-Augmented Generation}
\label{sec:rel_cag}

To mitigate high prefill latency in RAG, recent optimizations leverage Cache-Augmented Generation (CAG) to precompute and reuse document KV caches~\citep{gim2024prompt}.
\citet{chan2025don} pioneered this by preloading the entire corpus into the context window to bypass online retrieval, at the cost of severe memory bloat from attending to the entire corpus.
\textbf{\standardCAG{}} instead bypasses online encoding by caching full retrieved chunks offline via $\mathrm{KV}_{\text{StandardCAG}} = C_{\text{ctx}}^{\text{chunk}} \oplus \mathrm{KV}_{\mathcal{M}}(q; C_{\text{ctx}}^{\text{chunk}})$, where $C_{\text{ctx}}^{\text{chunk}} = C_p \oplus \text{Rot}(C_{c_1}; \Delta_1) \oplus \dots \oplus \text{Rot}(C_{c_{k_c}}; \Delta_{k_c})$ represents the prefix context cache assembled from the full chunk representations, loaded into GPU memory at inference time.
TurboRAG~\citep{lu2025turborag} and Block-attention~\citep{ma2025block} instantiate this paradigm, precomputing and reusing chunk-level KV caches to isolate cross-chunk attention and slash TTFT.
As the two are essentially identical in design, we let \textbf{TurboRAG} (Figure~\ref{fig:comparison}(c)) represent this paradigm throughout.

Building on \standardCAG{}'s cached full-chunk representations, \ours{} (Figure~\ref{fig:comparison}(e)) compiles the context via an optimized cache-slicing pipeline.
Like \standardCAG{}, it encodes each chunk independently offline, but at the finer granularity of nuggets rather than full chunks, achieving a compact memory footprint that reduces latency while fully preserving \standardCAG{}'s document-level semantic grounding.

A separate line of work restores cross-chunk interactions by partially recomputing cached KV states online.
\textbf{CacheBlend}~\citep{yao2025cacheblend} (Figure~\ref{fig:comparison}(b)), which we also compare against, reuses precomputed KV caches while selectively recomputing a small subset of tokens to restore cross-attention with preceding context.
\textbf{KVLink}~\citep{yang2025kvlink} (Figure~\ref{fig:comparison}(d)) instead inserts trainable link tokens into each chunk, whose KV states are computed by attending over the previous chunks to restore cross-chunk self-attention.

\begin{table}[htbp]
\centering
\scriptsize
\setlength{\tabcolsep}{4pt}
\begin{tabular}{lllll}
\toprule
\textbf{System} & \textbf{Retrieval} & \textbf{KV} & \textbf{Encoding} & \textbf{Training} \\
	&  \textbf{Unit} & \textbf{Reuse} & \textbf{Context} &  \textbf{Required} \\
\midrule
\standardRAG{}  & Chunk  & No  & All chunks              & No  \\
CacheBlend      & Chunk  & Yes & Selective & No  \\
TurboRAG        & Chunk  & Yes & Each chunk              & Yes \\
KVLink          & Chunk  & Yes & Each chunk + link       & Yes \\
\ours{}         & Nugget & Yes & Each chunk              & Yes \\
\bottomrule
\end{tabular}
\caption{CoinRAG vs. related RAG baselines.}
\label{tab:comparison}
\end{table}

\subsection{Nugget-based RAG}
\label{sec:rel_nugget}

The nugget concept~\citep{2003nugget} was proposed for IR/QA system evaluation, where a human assessor enumerates the essential information nuggets that a good response should contain.
AutoNuggetizer~\citep{pradeep2025great} automated this process using LLMs, which GINGER~\citep{lajewska2025ginger} and Crucible~\citep{dietz2026incorporating} extend to nugget-augmented generation by dynamically constructing nuggets from retrieved documents to guide inference.

As listed in Table~\ref{tab:nugget}, \ours{} differs from these studies in both when and how nuggets are formed.
\ours{} extracts nuggets offline and query-independently, becoming query-specific only after online two-stage retrieval, and each nugget is a document span that recovers the broader context of its source chunk during inference.
GINGER and Crucible instead construct nuggets online per query with an LLM, each standing alone with no surrounding context.
Because they do not study KV cache precomputation and incur high prefill latency from repeated expensive LLM calls during query processing, we exclude them from our empirical comparison, which centers on KV cache reuse systems.

\begin{table}[htbp]
\centering
\scriptsize
\begin{tabular}{lcc}
\toprule
\textbf{Property} & \textbf{GINGER/Crucible} & \textbf{\ours{}} \\
\midrule
KV Cache Precomp. & No & Yes \\
Nugget Generation & Online & Offline \\
Nugget Vocabulary & Any text & Document span \\
Nugget Context & Context-free & Contextual \\
Latency & High & Low \\
\bottomrule
\end{tabular}
\caption{\ours{} vs. prior nugget-based RAG.}
\label{tab:nugget}
\end{table}

\section{Experiments}
\label{sec:experiments}

\subsection{Experimental Setup}

\paragraph{Baselines.}
We compare \ours{} and \ours{}+CacheBlend (\S\ref{sec:cacheblend_integration}) with \standardRAG{}, TurboRAG, CacheBlend, and KVLink, formalized in Section~\ref{sec:comparison}. Block-attention is omitted since TurboRAG represents it, given their near-identical chunk-level KV cache reuse design.
We replicate TurboRAG and KVLink, training them with the same data and hyperparameters as \ours{} for a fair comparison, though their training configurations necessarily differ given each method's distinct mechanism (Appendix~\ref{app:hyperparameters}).
\standardRAG{} and CacheBlend are training-free and use the original backbone.

\paragraph{Metrics.}
We adopt the token-level F1-score as our accuracy metric, which reliably reflects semantic correctness since our inference prompt (Appendix~\ref{app:generation_prompt}) constrains the model to synthesize concise answers without verbose explanations.
For efficiency, we focus on Time-to-First-Token (TTFT) latency (ms), the time to construct the prefix context and process the query during the prefill stage, which dominates overall inference time given our short generated answers.
TTFT is measured end to end from query arrival, including retrieval and cache loading (Appendix~\ref{sect:setupdetails}).
We additionally report the prefill length in tokens, a direct proxy for KV cache memory usage (56\,KB per token for our backbone), which often dominates GPU memory in long-context inference.

\begin{figure*}[t]
\centering
\includegraphics[width=0.85\linewidth]{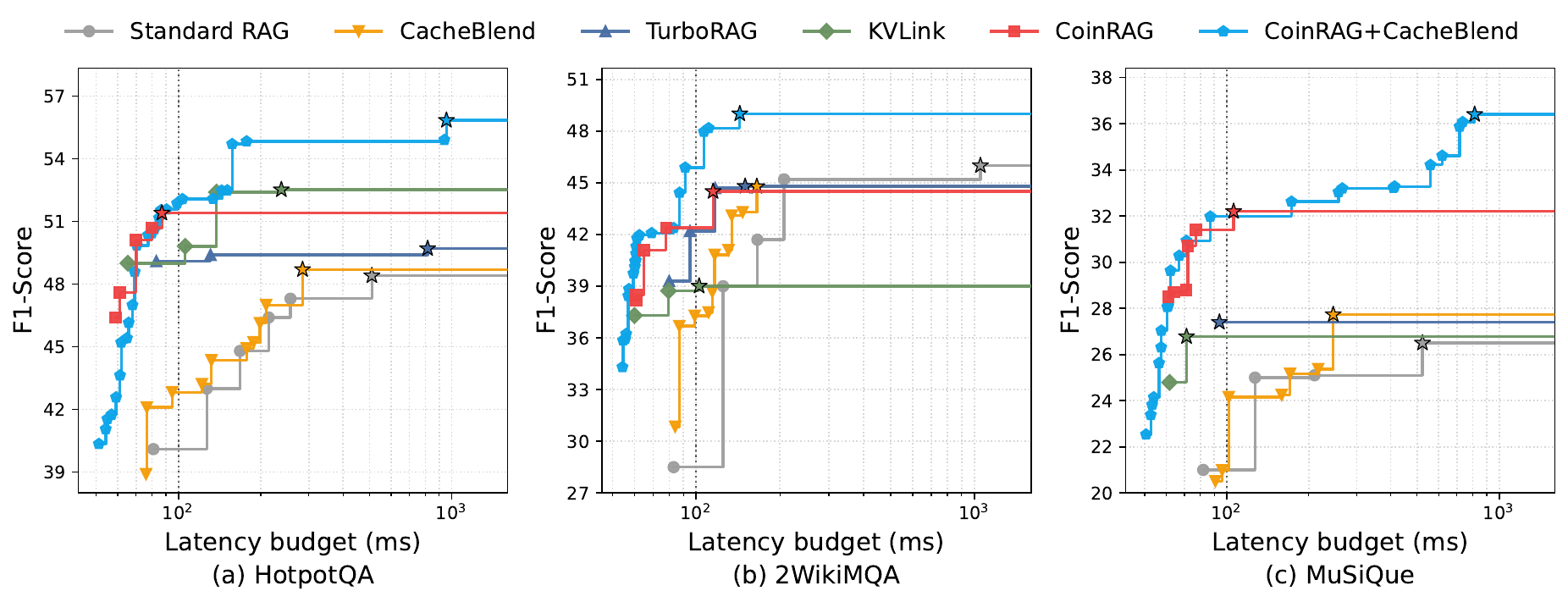}
\caption{
\label{fig:latency}
Pareto frontiers: accuracy vs. latency budget. 
The x-axis denotes 
Time-to-First-Token (TTFT) P99 latency budget on a log scale, while the y-axis shows F1-score on answer quality.
Star markers ($\star$) indicate the peak F1-score achieved by each method with no latency limit.
A vertical dashed line denotes the 100 ms latency budget.
}
\end{figure*}

\begin{table*}[t]
\centering
\resizebox{\textwidth}{!}{%
\setlength{\tabcolsep}{1pt}
\begin{tabular}{l|ccccc|ccccc|ccccc|cccc}
\toprule
\multirow{2}{*}{\textbf{Method}} & \multicolumn{5}{c|}{\textbf{HotpotQA}} & \multicolumn{5}{c|}{\textbf{2WikiMQA}} & \multicolumn{5}{c|}{\textbf{MuSiQue}} & \multicolumn{4}{c}{\textbf{Average}} \\
 & \textbf{\#} & \textbf{F1} $\uparrow$ & \textbf{TTFT} $\downarrow$ & \textbf{p99} $\downarrow$ & \textbf{Len} $\downarrow$ & \textbf{\#} & \textbf{F1} $\uparrow$ & \textbf{TTFT} $\downarrow$ & \textbf{p99} $\downarrow$ & \textbf{Len} $\downarrow$ & \textbf{\#} & \textbf{F1} $\uparrow$ & \textbf{TTFT} $\downarrow$ & \textbf{p99} $\downarrow$ & \textbf{Len} $\downarrow$ & \textbf{F1} $\uparrow$ & \textbf{TTFT} $\downarrow$ & \textbf{p99} $\downarrow$ & \textbf{Len} $\downarrow$ \\
\midrule
\multicolumn{20}{c}{\textbf{P99 latency $\le$ 100 ms}} \\
\midrule
\standardRAG{} & 1 & 40.1 & 74 & 81 & 741 & 1 & 28.5 & 71 & 83 & 684 & 1 & 21.0 & 76 & 82 & 743 & 29.9 & 74 & 82 & 723 \\
CacheBlend & 2 & 42.8 & 80 & 95 & 1175 & 3 & 37.3 & 80 & 99 & 1415 & 3 & 21.0 & 86 & 96 & 1576 & 33.7 & 82 & 97 & 1389 \\
TurboRAG & 1 & 49.1 & 71 & 83 & 752 & 2 & 42.2 & 80 & 95 & 1059 & 1 & 27.4 & 73 & 94 & 754 & 39.6 & 75 & 91 & 855 \\
KVLink & 1 & 49.0 & 57 & 65 & 749 & 3 & 38.7 & 70 & 80 & 1433 & 2 & 26.8 & 67 & 71 & 1170 & 38.2 & 65 & 72 & 1117 \\
\ours{} & 1, 30 & \underline{51.4} & 72 & 87 & 618 & 3, 3 & \underline{42.4} & 61 & 78 & 389 & 4, 3 & \underline{31.4} & 62 & 77 & 387 & \underline{41.7} & 65 & 81 & 465 \\
\ours{}+CacheBlend & 2, 50, .3 & \textbf{51.9} & 74 & 99 & 927 & 2, 50, .5 & \textbf{45.9} & 72 & 91 & 835 & 4, 20, .1 & \textbf{32.0} & 67 & 87 & 742 & \textbf{43.2} & 71 & 92 & 835 \\
\midrule
\multicolumn{20}{c}{\textbf{No latency limit constraint}} \\
\midrule
\standardRAG{} & 10 & 48.4 & 436 & 510 & 4687 & 20 & \underline{46.0} & 820 & 1051 & 8241 & 10 & 26.5 & 429 & 536 & 4635 & 40.3 & 562 & 699 & 5854 \\
CacheBlend & 5 & 48.7 & 222 & 285 & 2484 & 5 & 44.8 & 130 & 165 & 2221 & 5 & 27.7 & 208 & 246 & 2442 & 40.4 & 187 & 232 & 2382 \\
TurboRAG & 20 & 49.7 & 439 & 816 & 9262 & 4 & 44.8 & 116 & 150 & 1849 & 1 & 27.4 & 73 & 94 & 754 & 40.6 & 209 & 353 & 3955 \\
KVLink & 20 & \underline{52.5} & 189 & 238 & 9175 & 5 & 39.0 & 87 & 102 & 2251 & 2 & 26.8 & 67 & 71 & 1170 & 39.4 & 114 & 137 & 4199 \\
\ours{} & 1, 30 & 51.4 & 72 & 87 & 618 & 3, 20 & 44.5 & 96 & 115 & 695 & 5, 5 & \underline{32.2} & 76 & 106 & 427 & \underline{42.7} & 81 & 103 & 580 \\
\ours{}+CacheBlend & 20, 200, .9 & \textbf{55.8} & 414 & 956 & 4311 & 3, 50, .7 & \textbf{49.0} & 97 & 143 & 1048 & 30, 150, 1 & \textbf{36.4} & 367 & 813 & 3300 & \textbf{47.1} & 293 & 637 & 2886 \\
\bottomrule
\end{tabular}
}
\caption{
F1 score of the best configuration of each baseline under two latency budgets. 
The \# column denotes the chunk count $k_c$, the nugget composition $k_c, k$, or $k_c, k, r$ with the recomputation ratio $r$, TTFT is prefill latency in milliseconds, and Len is the active prefix context length in tokens. \textbf{Bold} and \underline{underline} mark the highest and second-highest F1 across methods within each budget and dataset (or average).}
\label{tab:latency}
\end{table*}

\paragraph{Evaluation Datasets.}
We conduct experiments on three multi-document question answering benchmarks from LongBench~\citep{bai2024longbench} that demand cross-document aggregation and reasoning: HotpotQA~\citep{yang2018hotpotqa}, 2WikiMQA~\citep{ho2020constructing}, and MuSiQue~\citep{trivedi2022musique}.
Each dataset provides a task-specific document corpus for retrieval.

\paragraph{Other settings.}
During offline time, we employ GPT-4o-mini to extract nuggets from 512-token chunks.
Online retrieval and answer generation use \embed{} and \baseLLM{}, chosen for its native support for flexible position rotation via RoPE.
For each method, we sweep the number of retrieved chunks $k_c$ (or nuggets $k$ for \ours{}) to draw its accuracy--efficiency trade-off curve.
For CacheBlend and \ours{}+CacheBlend, we additionally sweep the recomputation ratio $r$.
Under each budget, each method reports its best-F1 configuration on the evaluation set, with deterministic greedy decoding.
We fine-tune \ours{} for one epoch over 277,280 training instances.
Appendix~\ref{sect:setupdetails} gives additional setup details.

\subsection{Main Results}
\label{sec:experimental_results}

Figure~\ref{fig:latency} shows Pareto frontiers, with F1 accuracy score on the y-axis and P99 TTFT latency budget on the x-axis, plotting each baseline's F1 score with its best configuration under a given budget.
As the budget increases, a baseline may reach its peak F1 score, after which its curve becomes flat.

Table~\ref{tab:latency} compares the baselines on 3 datasets under a P99 TTFT latency budget of 100 ms (top) and with no latency limit (bottom), reporting the F1 score, mean TTFT, P99 TTFT, and average input token length of each baseline's best configuration.
The last few columns list the average performance across the 3 datasets.

\begin{figure*}[t]
\centering
\includegraphics[width=0.85\linewidth]{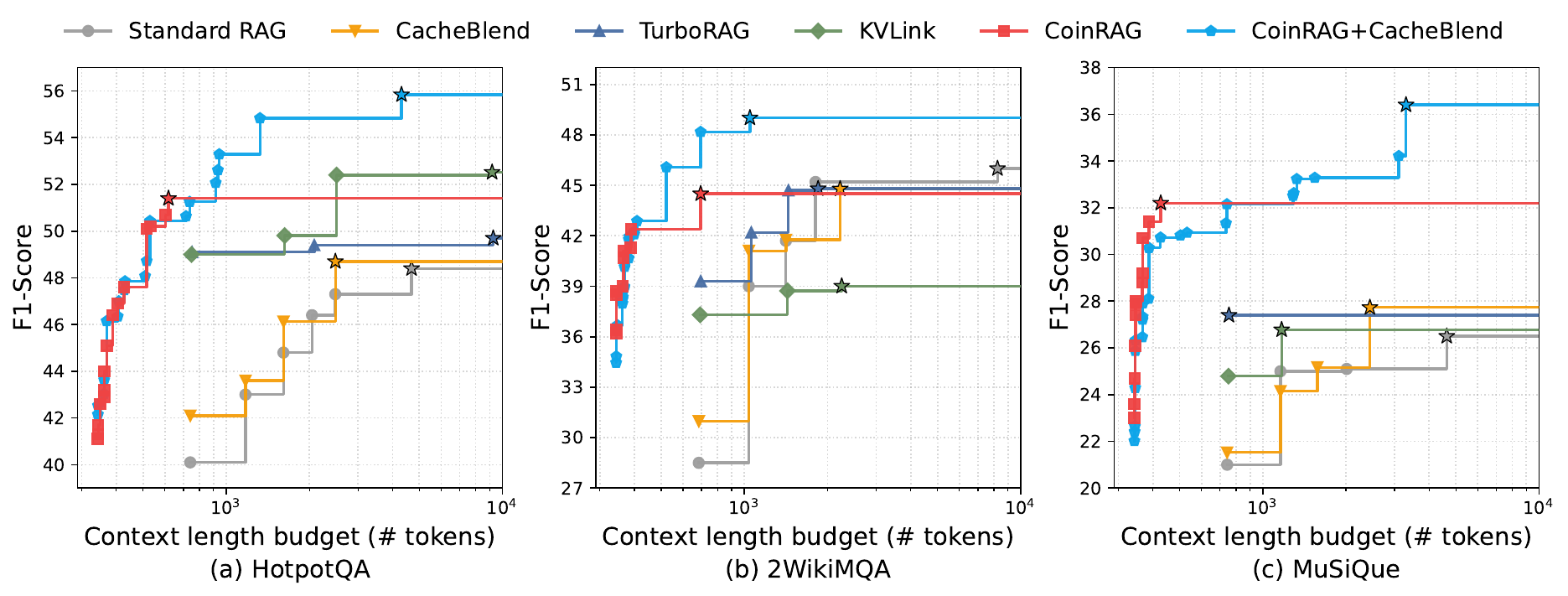}
\caption{
\label{fig:length}
Pareto frontiers: accuracy vs. length budget.
The x-axis denotes a prefill context length limit in token count on a log scale, while the y-axis shows F1-score on answer quality.
Star markers ($\star$) indicate the peak F1-score achieved by each method with no context length limit.
}
\end{figure*}

With a P99 100 ms latency constraint, \standardRAG{} is forced only to retrieve 1 document to meet the budget, while other baselines offer more flexibility with KV cache reuse.
In all 3 datasets, \ours{} outperforms \standardRAG{}, CacheBlend, TurboRAG, and KVLink, while TurboRAG is the strongest among the other baselines.
The F1 score of \ours{} is 5.3\% higher than TurboRAG averaged over 3 datasets, while the contextual token length of \ours{} is 1.84$\times$ shorter.
A paired bootstrap test over questions (Appendix~\ref{sect:setupdetails}) shows that the gain over \standardRAG{} is significant ($p<.001$).
The gain over TurboRAG is consistent across all 3 datasets but within the 95\% confidence interval, and the two are comparable on 2WikiMQA.

When relaxing the P99 latency budget to 116 ms, \ours{} still outperforms all other baselines in F1 score.
After that, \standardRAG{} and TurboRAG begin to perform comparably to \ours{} and gradually outperform it for 2WikiMQA, as shown in Figure~\ref{fig:latency}.
After around 160 ms, KVLink starts to catch up to and outperform \ours{} on HotpotQA.
This means cross-chunk interaction can contribute positively, but can also lead to more noise.
\ours{}+CacheBlend further improves F1 under the same budget, reaching an average of 43.2 at 100 ms (3.6\% over \ours{}, with $r \le 0.5$) and the highest F1 on all 3 datasets with no latency limit (47.1 on average).

When removing the latency limit entirely, TurboRAG remains the strongest other baseline, and \ours{} outperforms it in 2 out of 3 datasets with a 3-dataset average improvement of 5.2\% (42.7 vs. 40.6) and with a 6.8$\times$ shorter average length.
Although the slim KV cache representation in \ours{} can sometimes miss useful interactions across chunks, on average this design can still offset the loss with a larger gain by removing noise and unnecessary context that misleads or slows down LLM inference.

Figure~\ref{fig:length} shows Pareto frontiers analogous to Figure~\ref{fig:latency}, but with the prefill context length limit in tokens on the x-axis instead of TTFT.
Given the same length limit, \ours{} achieves a higher F1 score than the other baselines, aided by its nugget-level KV cache reuse, which yields a substantially more compact context representation.
Even with no length limit, \ours{}'s context length remains up to 10.1$\times$ shorter than \standardRAG{}'s, the longest among all baselines under this setting.
This active token count serves as a direct proxy for the live KV cache size, which dictates the runtime GPU memory footprint during inference.
Across both the latency and length axes, \ours{} establishes a stronger empirical Pareto frontier, maximizing answer quality while substantially reducing hardware resource usage.

\subsection{Ablation Studies}
\label{sec:ablation}

\begin{figure*}[t]
\centering
\includegraphics[width=0.85\linewidth]{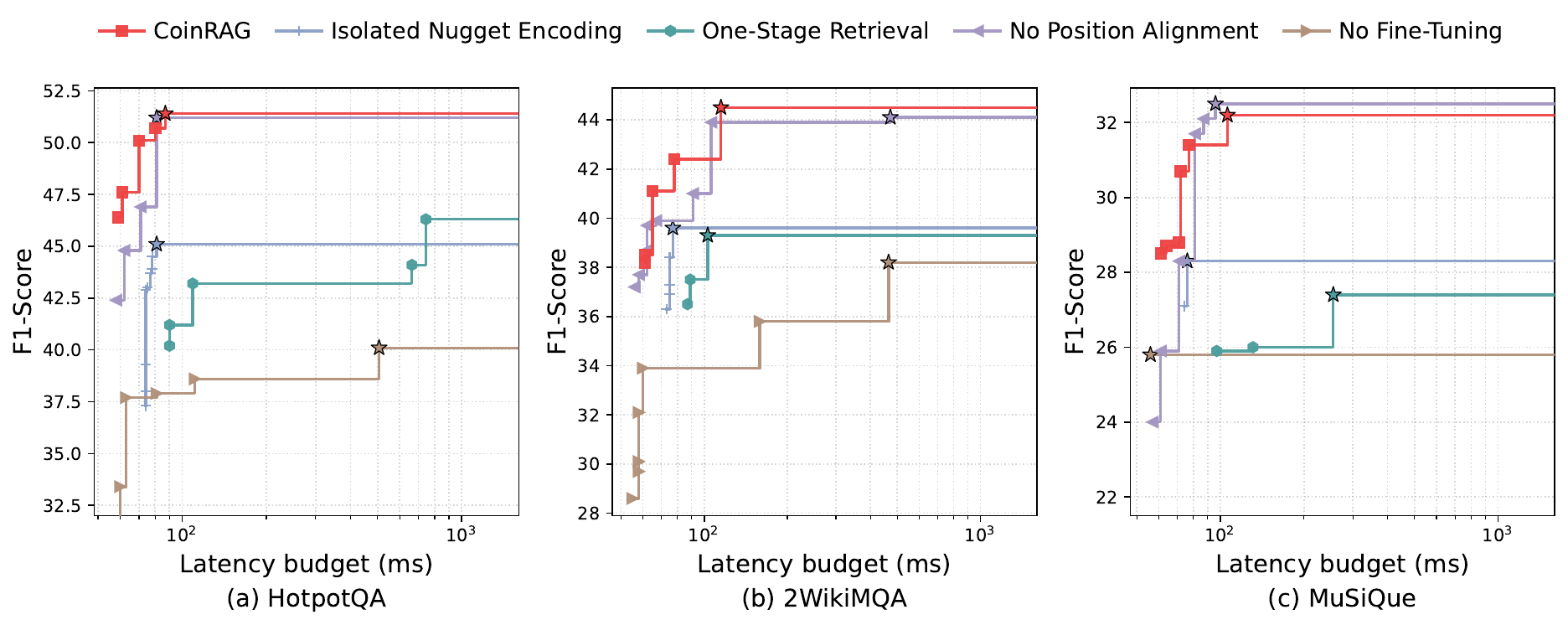}
\caption{
\label{fig:ablation}
F1-score versus TTFT P99 latency budget (ms) comparing \ours{} against four ablations, each removing one component: Isolated Nugget Encoding, One-Stage Retrieval, No Position Alignment, and No Fine-Tuning. Star markers ($\star$) indicate each configuration's peak F1-score.
}
\end{figure*}

Figure~\ref{fig:ablation} plots F1-score against TTFT for the following four ablation studies, each evaluated on HotpotQA, 2WikiMQA, and MuSiQue.

\paragraph{Contextualized vs. Isolated Nugget Encoding.}
We examine the effect of contextualized KV slicing (\S\ref{sec:cache_composition}) by comparing it against encoding each selected nugget span in isolation, using identical retrieval and token spans in both cases.
For \ours{}, encoding nuggets in isolation lowers peak F1-score by 6.3, 4.9, and 3.9 points on the 3 datasets.
This confirms that the cached KV representations carry contextual information beyond what the isolated nugget span alone conveys.

\paragraph{Two-Stage vs. One-Stage Retrieval.}
We compare \ours{}'s two-stage chunk-to-nugget retrieval (\S\ref{sec:nugget_online}) against retrieving nuggets directly from the full nugget collection in a single stage.
Two-stage retrieval reaches a higher peak F1-score on all 3 datasets, a relative gain of 9.6 to 17.5 percent over one-stage retrieval.
This F1 gain also comes with fewer selected nuggets at the peak configuration, which shortens the active context and lowers TTFT.
This confirms that narrowing the candidate pool to nuggets within already-retrieved chunks improves retrieval quality.

\paragraph{Position Alignment.}
We compare \ours{}'s position alignment (\S\ref{sec:cache_composition}) against applying no position alignment ($\Delta_i = 0$), which leaves each nugget at its original position within its own source chunk.
Position alignment is most beneficial under a tight 75 ms P99 budget, improving F1 by 3.0--8.5\% across the 3 datasets.
From a 100 ms P99 budget onward, position alignment achieves a comparable F1 score to no position alignment across the 3 datasets.

\paragraph{Nugget-Aware Fine-Tuning.}
We compare \ours{} with and without nugget-aware fine-tuning (\S\ref{sec:finetuning}).
Nugget-aware fine-tuning improves peak F1-score by $+$11.3 points on HotpotQA, $+$6.3 points on 2WikiMQA, and $+$6.4 points on MuSiQue.
This confirms that nugget-aware fine-tuning bridges the training-inference alignment gap by calibrating the model against position embedding shifts induced by stitching together non-contiguous nugget spans.

\section{Conclusion}
\label{sec:conclusion}

We present \ours{}, a lightweight long-context RAG framework with a slim representation to deliver effective and interactive responses via contextual KV nugget cache reuse.
Key underlying technical contributions include: an offline extraction of essential text-span-based information nuggets,
nugget-aware finetuning, a two-stage online retrieval pipeline that selects query-specific nuggets, and
contextualized KV cache composition.  

Empirically, under a standard 100 ms P99 latency budget, \ours{} outperforms the F1 score of the best competitor, TurboRAG, by 4.7\%, 0.5\%, and 14.6\% for 3 datasets, with an average gain of 5.3\%.
Without a latency limit, \ours{} still outperforms the baselines on average for 3 datasets by 5.2\% F1 improvement or more, indicating that the gain by removing noise and unnecessary context by \ours{} outweighs the loss due to missing some interaction.
Position alignment during online cache composition is most beneficial when the P99 latency budget is tight (e.g., under 75 ms), and remains comparable to no position alignment beyond that budget.
In general, \ours{} establishes a new Pareto frontier for effective long-context RAG inference under an interactive TTFT SLA budget.

\section*{Limitations}
\label{sec:limitations}

While \ours{} has demonstrated a new Pareto frontier for effective long-context RAG under a standard low-latency SLA budget, we explicitly acknowledge several systemic constraints.
\begin{itemize}[leftmargin=*, topsep=2pt, itemsep=0pt]
\item 
Although pre-encoding text into static KV representations significantly accelerates online prefill runtimes, it introduces offline resource costs that scale with the corpus size, including disk storage overhead, which may call for hierarchical or tiered caching infrastructure for large corpora, and a non-trivial one-time nugget-aware fine-tuning cost.
We assume documents do not change frequently, and thus one-time or infrequent offline computing cost can be acceptable.
For more dynamic corpora, each chunk is cached independently, so updating a document requires re-extracting nuggets and re-encoding only its own chunks rather than the whole cache.
Appendix~\ref{app:infrastructure} describes offline cache construction, disk storage, selective loading, and the I/O setup in more detail.
\item 
The precomputed cache layer is inherently coupled to the specific model checkpoint.
Because the stored KV representations are tightly bound to the underlying transformer weights and positional embedding topologies, updating the backbone language model or altering its architecture forces a re-encoding of documents.
The cache size also grows with the number of layers and KV heads of the backbone (about 28\,MB per 512-token chunk for our 7B backbone), as it does for all KV cache reuse methods, while the prefill computation that cache reuse saves grows with the backbone as well.
Appendix~\ref{app:future_work} discusses the choice of backbone size under a tight latency budget.
\item 
The downstream answer quality remains bounded by the initial chunk-level and nugget-level retrieval recall.
If the retrieval pipeline fails to surface the canonical evidence spans, the downstream model cannot synthesize the correct answer, although integrating task-aware fine-tuned retrievers or cross-encoder rerankers represents a promising trajectory to mitigate this orthogonal bottleneck.
\item 
\ours{} composes cached representations at the granularity of individual chunks, so nuggets originating from different chunks do not attend to each other during encoding, a limitation shared with TurboRAG-style chunk-level caching.
While our evaluation has compared CacheBlend and KVLink, which address cross-chunk attention in KV cache reuse, Section~\ref{sec:cacheblend_integration} and Appendix~\ref{app:cross_chunk} further combine each of them with \ours{}.
Combining with CacheBlend's selective recomputation further improves F1 under the same latency budget, while combining with KVLink yields mixed results across the 3 datasets.
\item Our evaluation assumes that the online cache has a capacity limitation and thus does not consider inter-query KV-cache reuse when some consecutive queries may share the same retrieved documents.
Prior KV-cache reuse studies, including TurboRAG, KVLink, and CacheBlend, follow this assumption, and this setting is fair for comparing them.
With large online cache capacity, standard RAG could take advantage of document overlap among queries to reduce TTFT, while the F1 score of standard RAG with no latency-limit budget in Table~\ref{tab:latency} remains the same.  
While query-independent offline KV cache reuse can be more versatile, the impact evaluation on exploiting inter-query cache reuse requires a traffic dataset with a query appearance order, and this can be studied in the future.
Even without a traffic dataset, queries in each evaluation set retrieve from a shared corpus, and at $k_c=30$, 36--61\% of retrieval events land on a chunk already retrieved by an earlier query.
\item Our conclusions are drawn within the tested setting: one backbone (\baseLLM{}) for the main comparison, one nugget extractor, one retriever, one inference GPU with single-stream serving, and three English multi-hop QA benchmarks.
We report the KV cache footprint, which is exact for a given prefill length, while serving throughput under concurrent load is inferred from it rather than measured.
\end{itemize}
We discuss additional future work in Appendix~\ref{app:future_work}.


\bibliography{reference}

\clearpage

\appendix

\section{Future Work}
\label{app:future_work}

We evaluate \ours{} on three LongBench multi-hop, multi-document QA benchmarks (HotpotQA, 2WikiMQA, MuSiQue) because answering their queries requires synthesizing evidence scattered across multiple retrieved chunks, directly stressing the cross-chunk retrieval and cache reuse behavior that \ours{} targets, unlike single-document or single-hop QA benchmarks; this dataset choice also follows common practice across prior KV cache reuse works.
Our evaluation focuses on single-shot multi-document question answering. Extending \ours{} to genuinely conversational or repeated-query benchmarks
(e.g., QuAC~\citep{choi2018quac}, CoQA~\citep{reddy2019coqa}, Doc2Dial~\citep{feng2020doc2dial}, MultiDoc2Dial~\citep{feng2021multidoc2dial}), where retrieval happens 
multiple times over a conversation and often returns the same or overlapping document context across turns, would further validate 
the amortized benefit of offline cache reuse and broaden its applicability.

Our evaluation holds one latency budget and one retrieval configuration fixed across all queries within each dataset, so that methods are compared at matched serving cost.
In deployment, a harder multi-hop query may justify a larger budget than a simple factoid query.
Query-adaptive budget allocation is therefore a promising direction, in line with prior work that adapts retrieval or context to each query~\citep{asai2024self, trivedi2023interleaving, xu2024recomp, yoon2024compact, lim2025macrag}, and \ours{} is well suited to it.
The per-query nugget count $k$ is a natural control knob, which could be set by a retrieval score threshold or a query complexity predictor, and because each nugget adds only a short span to the prefix, the context and prefill cost can be adjusted in much finer increments than with whole chunks.

We focus on a 7B backbone because prior KV cache reuse methods that require training are evaluated at this scale: TurboRAG~\citep{lu2025turborag} uses Qwen2-7B-Instruct, the same backbone as ours, KVLink~\citep{yang2025kvlink} uses Llama models from 1B to 8B, and Block-attention~\citep{ma2025block} uses an 8B model.
Only CacheBlend~\citep{yao2025cacheblend} is evaluated at larger scales, as it requires no training.
Since \ours{}, TurboRAG, and KVLink are all fine-tuned, training cost bounds the backbone scale for all of them alike.
The mechanism of \ours{} itself makes no scale-specific assumption, since KV slicing and RoPE-based position alignment operate on the cache layout and positional encoding regardless of parameter count.
A study at larger scales, including the trained baselines, would further establish how the accuracy--latency trade-off of nugget-level cache reuse evolves with model size, as both prefill savings and per-chunk cache size grow with the backbone.
At 7B, \ours{} already delivers strong accuracy under the 100\,ms budget, and since a larger backbone raises both the prefill and the cache-loading cost of every context token, a moderately sized backbone can be a practical choice for tight interactive budgets without further I/O optimization.
At larger scales, the shorter prefix of \ours{} also saves more prefill computation per query.

Finally, prompt injection~\citep{greshake2023not} is a broader security concern for RAG systems in general, as retrieved documents may contain malicious content. 
Because cached nugget representations in \ours{} are computed offline, standard prompt-injection mitigations such as anomaly detectors and sanitizers could be applied at that stage without affecting online latency.
Nonetheless, answer generation may still attend to cached malicious content if an adversarial span is retrieved, and we leave a systematic study of nugget cache reuse's susceptibility to prompt injection to future work.

\section{Answer Generation Prompt}
\label{app:generation_prompt}
\begin{figure*}[ht]
\centering
\fbox{
\parbox{0.98\textwidth}{\small\itshape
\#\#\# Instruction:\\
You are an intelligent Open Domain QA system.\\
You will be given a [Question] and a list of candidate [Clues].\\
You must determine the correct answer based on your own internal knowledge, using the [Clues] as potential clues.\\

\textbf{\upshape Follow this logic strictly:}\\

\textbf{\upshape 1. Direct Match Priority}\\
- Examine all [Clues].\\
- If any clue contains the exact answer (exact span, alias, abbreviation, numerical value, or canonical entity), output it verbatim.\\

\textbf{\upshape 2. Retrieval-Guided Reasoning (Clues-First Inference)}\\
If no clue contains the exact answer, but:\\
- a clue includes key entities, partial spans, events, definitions, dates, or semantic cues relevant to the question,\\
then:\\
- Use \emph{only} these retrieved cues to infer the correct short answer.\\
The inference must stay anchored to the retrieved content, not free-form reasoning.\\
Example:\\
- If the clue is "The Big Apple" and the question is "Which city … ?", answer "New York City".\\

\textbf{\upshape 3. Minimal Fallback (Only When Retrieval Is Useless)}\\
- Only if none of the retrieved clues provide helpful information\\
(no match, no partial cue, no relevant entity),\\
- Then rely on your internal knowledge to generate the correct short answer.\\

\textbf{\upshape Output Rules}\\
Output only the short answer.\\
No explanations.\\
No reasoning.\\
Answer should be minimal (entity, number, date, phrase).\\

\#\#\# Input:\\
\textcolor{blue}{\{\(N\) clues, each prefixed by {\upshape\texttt{<|doc\_start|>}}\}}\\
Question: \textcolor{blue}{\{question\}}
}}
\caption{
\baseLLM{} answer generation prompt. 
The instruction block is computed once as the system prefix. 
The clues (chunks or nuggets) fill the cached K/V slot, and the question is appended. 
The model is constrained to emit a minimal short answer for lexical token-matching scoring.}
\label{fig:generation_prompt}
\end{figure*}

Figure~\ref{fig:generation_prompt} presents the full text of the prompt template injected by \ours{} into the base language model ($\mathcal{M}$) to generate a short answer from the retrieved nuggets.
The instruction block explicitly enforces a three-tier execution priority: 
(i)~if the correct answer is explicitly contained within the provided clues, it must be extracted verbatim, 
(ii)~if not explicitly stated, the model must deduce the answer through multi-hop reasoning strictly grounded in the retrieved context, and 
(iii)~the model may resort to its internal parametric knowledge only when the prior conditions are completely unfulfilled. 
This core instruction is pre-encoded only once as the static system prefix described in Section~\ref{sec:cache_composition}, whereas the dynamic nugget (clue) sequences are mapped directly to the precomputed full-context KV cache slots, with the target question appended at the very end of the prompt sequence.

\section{Offline Nugget Extraction: Prompt, Example, and Characteristics}
\label{app:nugget}

To operationalize the offline extraction framework of \ours{}, we leverage GPT-4o-mini to derive an initial list of nuggets via the structured layout detailed in Figure~\ref{fig:nugget_prompt}, which specifies the complete system prompt template and underlying constraints utilized throughout the offline processing phase.
The number of candidate nuggets $K$ proposed per passage (Algorithm~\ref{alg:nugget}, Line 1) is not a fixed setting but an outcome of the LLM's output for each passage, averaging approximately 7.1 candidates per passage across the training and evaluation corpora.

\begin{figure*}[ht]
\centering
\fbox{\parbox{0.98\textwidth}{\small\itshape
You are an information extraction system for Open-Domain Question Answering.\\

Given a passage, extract all spans that could serve as evidence or direct answers to factual questions.\\

\#\#\# Rules:\\
1. Extract spans \textbf{\upshape EXACTLY} as they appear in the original text. Do NOT paraphrase, reorder, or modify any words.\\
2. A span can be any length — a clause, a full sentence, or multiple sentences — as long as it captures a single coherent fact. Choose the \textbf{\upshape minimal} span that makes the fact understandable on its own.\\
3. Focus on spans containing:\\
\mbox{}\ \ \ - Named entities with factual context (e.g., birth, death, roles, locations)\\
\mbox{}\ \ \ - Dates, quantities, measurements, statistics\\
\mbox{}\ \ \ - Events and actions (who did what, when, where)\\
\mbox{}\ \ \ - Relationships between entities (e.g., "X is the father of Y")\\
\mbox{}\ \ \ - Definitions or descriptions of concepts\\
4. Skip spans that are:\\
\mbox{}\ \ \ - Purely transitional or connective (e.g., "However, this was not the case.")\\
\mbox{}\ \ \ - Redundant (already covered by another extracted span)\\
\mbox{}\ \ \ - Opinions or speculations without factual grounding\\
5. Spans may partially overlap if they capture different facts.\\

\#\#\# Output Format:\\
Return a JSON array of strings. Each string must be an exact substring of the original passage.\\

\textbf{\upshape Example:}\\{}
["Albert Einstein was born on March 14, 1879, in Ulm, Germany.", "the Nobel Prize in Physics in 1921", "his explanation of the photoelectric effect"]
}}
\caption{Full system prompt used for GPT-4o-mini nugget extraction. The user content follows the format \texttt{"Passage:$\backslash$n\{passage\}"}.}
\label{fig:nugget_prompt}
\end{figure*}

Figure~\ref{fig:example} provides an example of step-by-step extraction executed on an actual LongBench text segment from the HotpotQA subset, specifically mapping the transition from a raw source paragraph to structured factual nuggets using a sample passage (\texttt{passage\_idx=7}, ``Grania: She-King of the Irish Seas'').

\begin{figure*}[t]
\centering
\fbox{\parbox{0.98\textwidth}{\small
\textbf{Input passage} (longbench-hotpotqa, passage\_idx=7):
\begin{quote}\itshape
Passage 7:\\
Grania: She-King of the Irish Seas\\
Grania: She-King of the Irish Seas is a 1986 historical fiction novel about \textcolor{blue}{Grace O'Malley (Irish: Gr\'ainne N\'i Mh\'aille)}, the so-called Sea Queen of Connemara, by American-born Irish author Morgan Llywelyn. Llywelyn's novel is a heavily fictionalized account of O'Malley's life, with the author having created characters as needed for the plot of the story. The novel was the basis for the 2007 Broadway musical The Pirate Queen.
\end{quote}

\medskip
\textbf{Step 1.}\quad $\mathcal{E}_{\text{LLM}}(p) = \{\tilde{n}_1, \ldots, \tilde{n}_6\}$
\begin{enumerate}[leftmargin=1.8em,topsep=2pt,itemsep=0pt]
  \item \texttt{Grania: She-King of the Irish Seas is a 1986 historical fiction novel}
  \item \texttt{\textcolor{blue}{Grace O'Malley (Irish: Grainne Ni Mhaille)}}
  \item \texttt{the so-called Sea Queen of Connemara}
  \item \texttt{by American-born Irish author Morgan Llywelyn}
  \item \texttt{Llywelyn's novel is a heavily fictionalized account of O'Malley's life}
  \item \texttt{The novel was the basis for the 2007 Broadway musical The Pirate Queen}
\end{enumerate}

\medskip
\textbf{Step 2.}\quad Stage A partition
\begin{description}[leftmargin=2em, topsep=2pt, itemsep=1pt]
  \item[Exact substring of $p$ ($n_i \gets \tilde{n}_i$):] $\tilde{n}_1, \tilde{n}_3, \tilde{n}_4, \tilde{n}_5, \tilde{n}_6$
  \item[Non-exact ($\rightarrow$ Stage B):] $\tilde{n}_2$
\end{description}

\medskip
\textbf{Step 3.}\quad Fuzzy recovery for $\tilde{n}_2$

\smallskip
\hspace*{1em} LLM candidate $\tilde{n}_2$:\\
\hspace*{2em} \textit{Grace O'Malley (Irish: Grainne Ni Mhaille)}

\smallskip
\hspace*{1em} $\downarrow$ \ \ best matching window in $p$ (fuzzy matching score $\mathrm{sim}(s^{\ast}, \tilde{n}_2) = 0.953$):

\smallskip
\hspace*{1em} \textcolor{blue}{$s^{\ast}$ (verbatim excerpt from $p$, highlighted in blue above)}:\\
\hspace*{2em} \textit{\textcolor{blue}{Grace O'Malley (Irish: Gr\'ainne N\'i Mh\'aille)}}

\medskip
\textbf{Step 4.}\quad Final $\mathcal{N}(p)$
\begin{enumerate}[leftmargin=1.8em, topsep=2pt, itemsep=0pt]
  \item \texttt{Grania: She-King of the Irish Seas is a 1986 historical fiction novel} \hfill \textit{exact}
  \item \texttt{\textcolor{blue}{Grace O'Malley (Irish: Gr\'ainne N\'i Mh\'aille)}} \hfill \textit{fuzzy}
  \item \texttt{the so-called Sea Queen of Connemara} \hfill \textit{exact}
  \item \texttt{by American-born Irish author Morgan Llywelyn} \hfill \textit{exact}
  \item \texttt{Llywelyn's novel is a heavily fictionalized account of O'Malley's life} \hfill \textit{exact}
  \item \texttt{The novel was the basis for the 2007 Broadway musical The Pirate Queen} \hfill \textit{exact}
\end{enumerate}
}}
\caption{A step-by-step example of the nugget extraction algorithm of \ours{} in Algorithm~\ref{alg:nugget}. The region in the passage targeted by fuzzy recovery is highlighted in \textcolor{blue}{blue}. All extracted nuggets are sub-sentence spans (clauses / phrases) rather than full sentences.}
\label{fig:example}
\end{figure*}

To evaluate the operational characteristics and density of the extracted text segments, we compile empirical statistics for both the training splits (IRCoT~\citep{trivedi2023interleaving}) and the LongBench evaluation splits of each dataset in Table~\ref{tab:dataset_stats}, covering query, passage, and nugget-level statistics.
The empirical results reveal that our offline pipeline consistently compresses diffuse document contexts into compact, atomic nuggets across both splits: mean nugget length is far shorter than mean passage length, while the share of overlapping adjacent nugget spans (Ovlp\%) remains low across all 3 datasets.
This confirms that \ours{}'s offline extraction pipeline behaves consistently from the smaller evaluation splits to the substantially larger training splits, and that \ours{} can drastically reduce context length without sacrificing the underlying semantic grounding necessary for accurate QA synthesis.

\begin{table*}[t]
\centering
\resizebox{\textwidth}{!}{%
\begin{tabular}{ll rr rrr rrrrr}
\toprule
\multirow{2}{*}{\textbf{Dataset}} & \multirow{2}{*}{\textbf{Split}} & \multicolumn{2}{c}{\textbf{Query}} & \multicolumn{3}{c}{\textbf{Passage}} & \multicolumn{5}{c}{\textbf{Nugget}} \\
\cmidrule(lr){3-4}\cmidrule(lr){5-7}\cmidrule(lr){8-12}
 & & \textbf{\#Q} & \textbf{Words} & \textbf{\#Psg} & \textbf{Psg/Q} & \textbf{Words} & \textbf{\#Nug} & \textbf{Nug/Q} & \textbf{Nug/Psg} & \textbf{Words} & \textbf{Ovlp\%} \\
\midrule
\multirow{2}{*}{\textbf{HotpotQA}} & \textbf{Train} & 91{,}010 & 17.9 & 858{,}618 & 9.4 & 94 & 6{,}662{,}421 & 73.2 & 7.8 & 10.0 & 4.1 \\
 & \textbf{Eval} & 200 & 15.2 & 1{,}719 & 8.6 & 1060 & 101{,}540 & 509.0 & 59.1 & 13.0 & 5.0 \\
\midrule
\multirow{2}{*}{\textbf{2WikiMQA}} & \textbf{Train} & 166{,}174 & 12.7 & 1{,}295{,}924 & 7.8 & 66 & 7{,}921{,}522 & 47.7 & 6.1 & 9.1 & 3.6 \\
 & \textbf{Eval} & 200 & 11.7 & 1{,}783 & 9.9 & 504 & 56{,}176 & 305.9 & 31.5 & 11.3 & 3.9 \\
\midrule
\multirow{2}{*}{\textbf{MuSiQue}} & \textbf{Train} & 20{,}096 & 16.1 & 392{,}913 & 19.6 & 83 & 2{,}647{,}463 & 131.7 & 6.7 & 9.8 & 4.5 \\
 & \textbf{Eval} & 200 & 15.6 & 2{,}179 & 11.1 & 1001 & 119{,}061 & 611.9 & 54.6 & 13.1 & 5.0 \\
\bottomrule
\end{tabular}}
\caption{Dataset statistics on the training set (IRCoT~\citep{trivedi2023interleaving}) and evaluation set (LongBench~\citep{bai2024longbench}). \textbf{\#Q}: training examples/test questions; \textbf{\#Psg}, \textbf{\#Nug}: total passages/nuggets (for train these are summed over the training examples, so they include cross-example repetition; for eval they are the unique test passages and their total nuggets); \textbf{Words}: mean length in words; \textbf{Psg/Q}: passages per question; \textbf{Nug/Q}: nuggets per training example/test question; \textbf{Nug/Psg}: nuggets per passage; \textbf{Ovlp\%}: share of adjacent nugget pairs whose spans overlap within a passage.}
\label{tab:dataset_stats}
\end{table*}

\section{Steps of Online Inference: Example}
\label{app:inference}

To illustrate the end-to-end execution flow of \ours{}, Figure~\ref{fig:inference} gives a step-by-step trace of answering a multi-hop question from the LongBench HotpotQA evaluation suite.
The entire operational sequence demonstrates how our framework bypasses online text re-encoding by dynamically assembling fine-grained cached segments into a seamless decoding prefix.

As shown in Figure~\ref{fig:inference}, the online processing pipeline operates through four main stages:
First, a dense embedding model retrieves the top-$k_c$ candidate text chunks from the large external corpus. 
Second, fine-grained information nuggets are retrieved exclusively within this pre-selected chunk pool to drop uninformative tokens while maintaining factual relevance. 
Third, the corresponding KV representations of the selected nuggets are directly fetched from the precomputed document cache and stitched together using our position alignment strategy to construct the final prompt prefix. 
Finally, the language model processes the online query conditioned on this optimized context cache to synthesize the correct answer.
\begin{figure*}[htbp]
\centering
\fbox{\parbox{0.98\textwidth}{\small\raggedright
\textbf{Input question} (longbench-hotpotqa, multi-hop):
\begin{quote}\itshape
Who wrote the novel that the 2007 Broadway musical The Pirate Queen was based on?
\end{quote}

\textbf{Step 1.}\quad Stage-1 chunk retrieval (\embed, top-$k_c=4$)
\begin{enumerate}[leftmargin=2.4em, topsep=2pt, itemsep=2pt]
  \item \textbf{[0.609]} \texttt{chunk\_2403} \hfill \textit{``Grania: She-King of the Irish Seas''} \\
        \textit{Grania: She-King of the Irish Seas is a 1986 historical fiction novel about Grace O'Malley (Irish: Gr\'ainne N\'i Mh\'aille), the so-called Sea Queen of Connemara, by American-born Irish author Morgan Llywelyn. Llywelyn's novel is a heavily fictionalized account of O'Malley's life\ldots}
  \item \textbf{[0.474]} \texttt{chunk\_3511} \hfill \textit{``Buxtehude Bull'' (German YA novel award list)} \\
        \textit{\ldots 2008: Markus Zusak, Die B\"ucherdiebin (The Book Thief); 2009: Suzanne Collins, Die Tribute von Panem -- T\"odliche Spiele (The Hunger Games); 2010: Susan Beth Pfeffer, Die Welt, wie wir sie kannten; 2011: Lauren Oliver, Delirium; 2012: John Green, Das Schicksal\ldots}
  \item \textbf{[0.447]} \texttt{chunk\_4862} \hfill \textit{``Gordon Dahlquist'' (playwright/novelist)} \\
        \textit{Gordon Dahlquist is an American playwright and novelist. A native of the Pacific Northwest, Dahlquist has lived and worked in New York City since 1988. His plays, which include Messalina and Delirium Palace (both Garland Playwriting Award winners), have been performed in New York and Los Angeles\ldots}
  \item \textbf{[0.445]} \texttt{chunk\_2379} \hfill \textit{``Robert Jordan'' (epic fantasy author)} \\
        \textit{James Oliver Rigney Jr. (October 17, 1948 -- September 16, 2007), better known by his pen name Robert Jordan, was an American author of epic fantasy. He is known best for his series The Wheel of Time\ldots}
\end{enumerate}

\textbf{Step 2.}\quad Stage-2 nugget retrieval within Stage-1 pool (top-$k=5$). \textcolor{blue}{Blue} = supporting; black = distractor.
\begin{enumerate}[leftmargin=2.4em, topsep=2pt, itemsep=2pt]
  \item \textbf{[0.883]} \textcolor{blue}{\texttt{nug\_38701}} \hfill \textit{from chunk\_2403 (Grania)} \\
        \textit{\textcolor{blue}{The novel was the basis for the 2007 Broadway musical The Pirate Queen}}
  \item \textbf{[0.516]} \texttt{nug\_38283} \hfill \textit{from chunk\_2379 (Robert Jordan)} \\
        \textit{an American author of epic fantasy}
  \item \textbf{[0.491]} \texttt{nug\_78735} \hfill \textit{from chunk\_4862 (Gordon Dahlquist)} \\
        \textit{Gordon Dahlquist is an American playwright and novelist.}
  \item \textbf{[0.437]} \texttt{nug\_38286} \hfill \textit{from chunk\_2379 (Robert Jordan)} \\
        \textit{one of several writers to have written original Conan the Barbarian novels}
  \item \textbf{[0.436]} \textcolor{blue}{\texttt{nug\_38699}} \hfill \textit{from chunk\_2403 (Grania)} \\
        \textit{\textcolor{blue}{by American-born Irish author Morgan Llywelyn}}
\end{enumerate}

\textbf{Step 3.}\quad RAG prompt fed to LLM (\textcolor{blue}{blue} = K/V loaded from pre-computed cache; black = newly prefilled at inference time). Each nugget is prefixed by a single \texttt{<|doc\_start|>} separator; no \texttt{<|doc\_end|>} is inserted between nuggets (it only appears once per chunk during KV cache pre-computation).

\smallskip
\begin{quote}
\ttfamily\footnotesize\raggedright
\textless|im\_start|\textgreater system\\
You are a helpful assistant. Use the context to answer the question concisely.\textless|im\_end|\textgreater\\
\textless|im\_start|\textgreater user\\
\textcolor{blue}{\textless|doc\_start|\textgreater The novel was the basis for the 2007 Broadway musical The Pirate Queen}\\
\textcolor{blue}{\textless|doc\_start|\textgreater an American author of epic fantasy}\\
\textcolor{blue}{\textless|doc\_start|\textgreater Gordon Dahlquist is an American playwright and novelist.}\\
\textcolor{blue}{\textless|doc\_start|\textgreater winning the Tony Award for Best Actor in a Musical}\\
\textcolor{blue}{\textless|doc\_start|\textgreater one of several writers to have written original Conan the Barbarian novels}\\
\textcolor{blue}{\textless|doc\_start|\textgreater by American-born Irish author Morgan Llywelyn}\\
Question: Who wrote the novel that the 2007 Broadway musical The Pirate Queen was based on?\textless|im\_end|\textgreater\\
\textless|im\_start|\textgreater assistant\textbackslash n
\end{quote}

\medskip
\textbf{Step 4.}\quad Output
\begin{quote}\itshape
\textcolor{blue}{Morgan Llywelyn}
\end{quote}
}}
\caption{Step-by-step illustration of the \ours{} inference pipeline (HotpotQA, $k_c=4$, $k=5$). \textit{All retrieval scores, chunk IDs, and nugget IDs come from an actual \embed run on the HotpotQA index.} The two supporting nuggets (\texttt{nug\_38701}, \texttt{nug\_38699}) both originate from the same chunk\_2403 (Grania passage), while ranks 2--5 in between are filled with distractors that share only surface-level overlap (``author / Broadway musical / Tony Award'').}
\label{fig:inference}
\end{figure*}

\section{Evaluation Setup Details}
\label{sect:setupdetails}

\paragraph{Chunk and Nugget Retrieval.}
During the offline phase, we employ GPT-4o-mini to extract atomic information nuggets from each chunk via structured prompting, accepting a fuzzy-matched candidate only if its similarity to the source span exceeds a threshold $\tau = 0.7$ (\S\ref{sec:nugget_offline}).
Additional details, including the extraction prompt, an example, and extraction statistics, are provided in Appendix~\ref{app:nugget}.
We use \embed{}~\citep{bge-m3} as our dense embedding model to fetch candidate text chunks of fixed size $512$ tokens and to score query-relevance of the nuggets extracted offline within the retrieved chunks.
For each method, we sweep retrieval top-$k$ to draw its accuracy--efficiency trade-off curve.
For chunk-based methods, we vary the number of retrieved coarse chunks within the range $k_c \in \{1, 2, 3, 4, 5, 10, 20, 30, 50\}$.
For the nugget-based method \ours{}, which uses two-stage retrieval, we first fetch candidate chunks using the identical $k_c$ range, and subsequently sweep the final selected nugget count within the range $k \in \{1, 2, 3, 4, 5, 10, 20, 30, 50, 100, 150, 200\}$.

\paragraph{KV Caching and Loading.}
Every chunk in the corpus is pre-encoded once into full-context KV representations and stored on disk as one file per chunk.
At inference time, \ours{} loads each retrieved chunk file, slices the token ranges of the selected nuggets on the CPU, and transfers only those slices to the GPU.
For CacheBlend, we sweep the recomputation ratio $r \in \{0.0, 0.1, 0.2, \ldots, 1.0\}$, in 0.1 intervals from 0 to 1.
Hardware infrastructure, cache storage and I/O, and offline cost figures are detailed in Appendix~\ref{app:infrastructure}.

\paragraph{Latency Measurement.}
TTFT is measured from query arrival, before retrieval, to the first-token logits at the end of prefill.
It covers query embedding, chunk and nugget retrieval, reading cached KV files from disk, slicing and position alignment, host-to-device transfer, and prefill, and excludes all offline work.
Before measurement, we warm up both the model forward path and the cache loading path.
Timing is wall-clock after \texttt{torch.cuda.synchronize()}, at batch size 1 with no concurrent requests, and every method runs on the same pipeline and hardware.
The OS page cache may retain chunk files across queries, while GPU-resident KV states are not kept between queries.
Each query is timed once, and P99 is computed over the 200 queries of each dataset.

\paragraph{Configuration Selection.}
For every method, the configuration reported under each budget is the one with the best F1 on the evaluation set among those meeting the budget, so Figures~\ref{fig:latency} and~\ref{fig:length} are achievable frontiers.
Decoding is greedy, so results are deterministic and do not depend on a random seed.
Under the 100\,ms budget, a paired bootstrap test over all 600 evaluation questions with 10,000 resamples gives an F1 gain of +11.9 (95\% CI [+8.5, +15.3], $p<.001$) over \standardRAG{} and +2.1 (95\% CI [$-$0.6, +4.9], $p=.13$) over TurboRAG.

\paragraph{Nugget-Aware Fine-tuning.}
We fine-tune the model for one epoch over 277,280 training instances from the training splits of HotpotQA, 2WikiMQA, and MuSiQue (per-dataset statistics in Table~\ref{tab:dataset_stats}), using an effective batch size of 16.
The hyperparameter configuration is detailed in Appendix~\ref{app:hyperparameters}.
Nugget extraction, cache construction, and nugget-aware fine-tuning are one-time costs per corpus and backbone checkpoint, not per query.

\paragraph{Generation and Prompting.}
The base language model $\mathcal{M}$ for cache slicing, composition, and downstream generation is \baseLLM~\citep{Qwen2}, chosen for its native support for flexible position rotation via RoPE.
We adopt an explicit QA-oriented prompt template that instructs the LLM to produce concise answers based on the retrieved evidence, avoiding verbose or descriptive outputs that complicate scoring.
The full prompt template of \ours{} is provided in Appendix~\ref{app:generation_prompt}.
Retrieved items are concatenated in ranked order from top-1 to top-$k$ when constructing the prefix context cache.

\section{Hardware Infrastructure and I/O Management}
\label{app:infrastructure}

To ensure the exact reproducibility of our efficiency evaluations, we specify the underlying computational environment and disk storage layout here. Throughout, we assume a static or infrequently updated document corpus, consistent with the fixed collections used in our evaluation, so the offline costs reported below are each incurred once rather than repeatedly. Even so, because each chunk is cached independently, supporting incremental corpus updates by re-encoding only the affected chunks would be a natural extension of this design.

\paragraph{Hardware.}
All experiments are conducted on a single node equipped with dual-socket AMD EPYC 9354 32-Core Processors (totaling 64 physical cores and 128 threads), 188\,GB of system memory, an NVIDIA RTX PRO 6000 (Blackwell Server Edition, 96\,GB VRAM), and an NVIDIA L40S (48\,GB VRAM). 
Model training is performed on the RTX PRO 6000, whereas all evaluation and latency benchmarks are executed on the L40S environment in \texttt{bfloat16} precision. 
The information nugget extraction pipeline is processed off-line via the GPT-4o-mini API with a concurrency limit of 100, bypassing local GPU computation. 
The precomputed KV caches are stored on a local NVMe SSD (Samsung PM9A3, 3.84\,TB, U.2, PCIe Gen4) and are loaded dynamically into GPU memory during training and inference.

\paragraph{KV Cache I/O Management.}
The primary inference overhead of \ours shifts from prefill computation to the disk I/O required for loading the retrieved chunk caches.
For the base model (\baseLLM; 28 layers, 4 KV heads, and a head dimension of 128), the KV cache occupies 56\,KB per token, or approximately 28\,MB for a single 512-token chunk in \texttt{bfloat16} precision.
Each chunk is stored as an independent file named \texttt{doc\_\{passage\_id\}\_\{chunk\_id\}.pt}, and these PyTorch tensors are loaded using a per-document thread pool (\texttt{--parallel\_mode per\_doc}).
Each retrieved chunk file is read in full, without memory-mapped or range reads, and the nugget token ranges are then sliced on the CPU.
Only the sliced KV states are transferred to the GPU, so slicing reduces host-to-device transfer, resident KV memory, and prefill computation, but not the bytes read from disk.
The reported TTFT therefore includes full-chunk file I/O, and storing or reading only nugget-covered positions would reduce it further.
Disk bytes read per query therefore scale with the number of retrieved chunks $k_c$, at about 28\,MB per chunk, as for TurboRAG with the same $k_c$.

\paragraph{Nugget Extraction.}
Extracting information nuggets from the training corpus (950,221 unique passages, yielding 6.08M nuggets at 99.9\% validity) via the GPT-4o-mini Batch API costs approximately \$265--325.
Extracting nuggets for the three evaluation corpora combined (5,681 passages, yielding 277K nuggets) costs under \$2.
Both are one-time costs incurred during corpus preparation rather than per query.

\paragraph{Offline Cache Construction.}
Building the complete chunk KV cache store for all three evaluation corpora combined takes under 30 minutes in total, since each chunk requires only a single forward pass, equivalent in FLOPs to one online prefill pass over that chunk's tokens.
Disk storage scales linearly with corpus size at approximately 27--28\,MB per chunk, totaling 146\,GB for HotpotQA, 75\,GB for 2WikiMQA, and 173\,GB for MuSiQue.
Nugget spans cover approximately 70\% of chunk tokens across the three evaluation corpora (after deduplicating positions shared by overlapping nuggets), implying that retaining only nugget-covered positions rather than the full chunk could reduce disk footprint by roughly 30\%.
Like nugget extraction, this is a one-time cost incurred during corpus preparation rather than per query.

\paragraph{Nugget-Aware Fine-Tuning.}
Nugget-aware fine-tuning trains \baseLLM{} for one epoch over 277,280 examples (17,330 optimizer steps at a gradient accumulation factor of 16), requiring approximately 140 GPU-hours on a single RTX PRO 6000 with a peak memory footprint of 75.8\,GB.
This is also a one-time cost, incurred once per backbone checkpoint rather than per query.

\section{Fine-Tuning Hyperparameters}
\label{app:hyperparameters}

The complete optimization configuration deployed for the nugget-aware fine-tuning pipeline is detailed in Table~\ref{tab:hyperparameters}. 
To ensure a fair and controlled comparison, we train the model under a fixed hyperparameter configuration. 
We employ gradient checkpointing and memory-efficient 8-bit Paged AdamW~\citep{dettmers2023qlora} to prevent out-of-memory (OOM) errors under strict sequence length constraints.

\paragraph{Training the Baselines.}
We train TurboRAG and KVLink on the same 277,280 training instances with the same hyperparameters as \ours{}.
All three methods minimize the cross-entropy loss on the answer tokens and differ only in how the prefix is constructed, as each mechanism requires.
TurboRAG uses independently encoded chunk KV caches with composite positions, in which each chunk keeps its own position IDs starting from 0~\citep{lu2025turborag}, KVLink uses chunk KV caches with 5 trainable link tokens per document, and \ours{} uses nugget KV slices with the order-preserving, contiguous position alignment of Section~\ref{sec:cache_composition}.
We do not adopt the hyperparameters reported in the original papers, since they were tuned for different backbones and training data, and a single shared recipe ensures that differences in accuracy reflect each mechanism rather than its training recipe.

\paragraph{Ablations without Fine-Tuning.}
To separate the cache representation from nugget-aware fine-tuning, we repeat two ablations of Section~\ref{sec:ablation} on the backbone without fine-tuning, whose full \ours{} configuration reaches a peak F1 score of 40.1, 38.2, and 25.8 on HotpotQA, 2WikiMQA, and MuSiQue.
Encoding nuggets in isolation lowers the peak F1 score by 6.1, 5.3, and 5.0 points, and one-stage retrieval lowers it by 5.0, 7.5, and 5.9 points.
These drops are similar to those with fine-tuning (Section~\ref{sec:ablation}), so the benefits of contextualized KV slicing and two-stage retrieval do not stem from fine-tuning alone.

\begin{table}[H]
  \centering
  \small
  \begin{tabular}{lc}
    \toprule
    \textbf{Hyperparameter} & \textbf{Value} \\
    \midrule
    Learning Rate            & $5\times10^{-6}$ \\
    LR Scheduler             & Cosine \\
    Warmup Ratio             & 0.03 \\
    Micro-Batch Size         & 1 \\
    Gradient Accumulation    & 16 \\
    Total Epochs             & 1 \\
    Optimizer                & Paged AdamW (8-bit) \\
    Precision                & \texttt{bfloat16} \\
    Gradient Checkpointing   & On \\
    Max Sequence Length      & 1024 \\
    Training Chunks ($k_c$)  & 5 \\
    Training Nuggets ($k$)   & 10 \\
    RAFT Noise Probability   & 0.15 \\
    \bottomrule
  \end{tabular}
  \caption{\label{tab:hyperparameters} Hyperparameter specifications for the nugget-aware fine-tuning pipeline.}
  
\end{table}

\section{Combining with Cross-Chunk Interaction}
\label{app:cross_chunk}

\begin{figure*}[t]
\centering
\includegraphics[width=\linewidth]{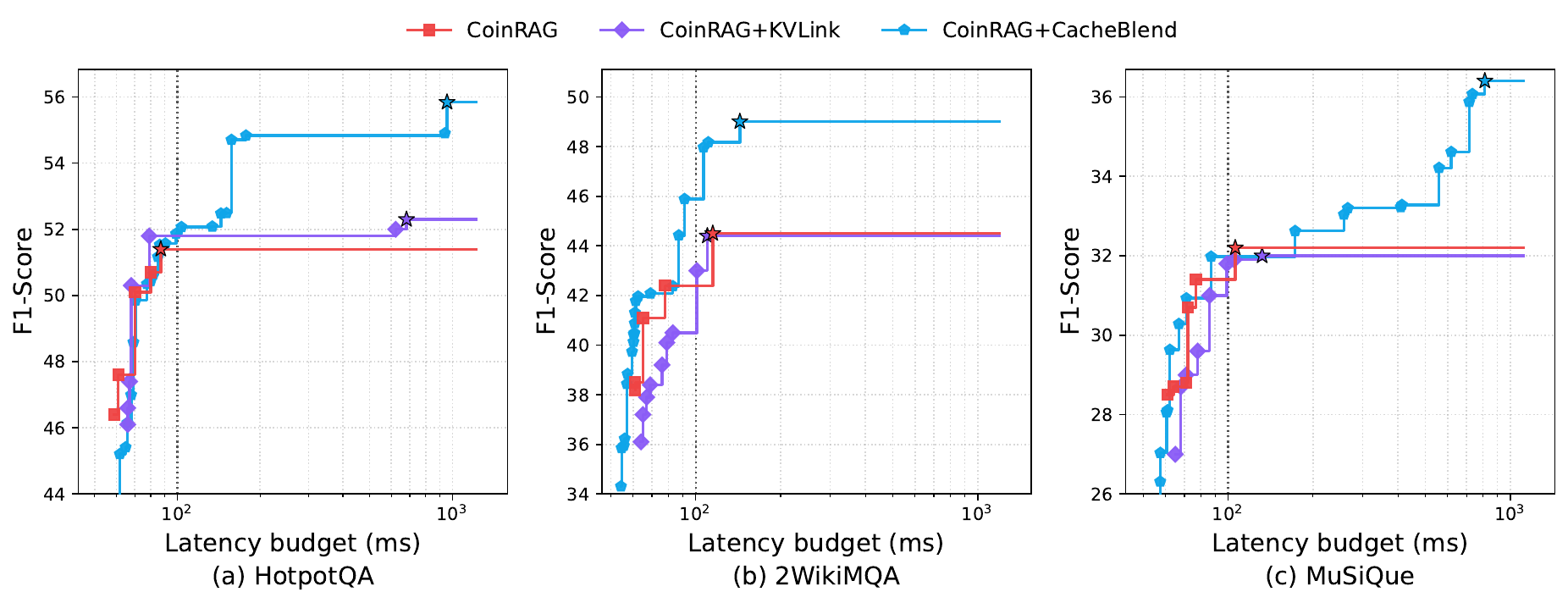}
\caption{
\label{fig:cross_chunk}
F1-score versus TTFT P99 latency budget (ms) for \ours{} alone and combined with KVLink or CacheBlend. Star markers ($\star$) indicate each configuration's peak F1-score, and a vertical dashed line denotes the 100 ms latency budget.
}
\end{figure*}

\ours{} encodes each chunk independently, so nuggets from different chunks do not attend to each other during encoding.
Section~\ref{sec:cacheblend_integration} integrates \ours{} with CacheBlend's selective recomputation, and we also combine it with KVLink's learned link tokens.
Figure~\ref{fig:cross_chunk} compares \ours{} alone with both combinations.
For \ours{}+KVLink, we insert KVLink's trainable link tokens after the nugget slices from each retrieved chunk, and compute them online by attending over all preceding slices.
The link-token embeddings and the backbone are fine-tuned jointly with the nugget-aware objective, using the same data and hyperparameters as \ours{}.
Combining with CacheBlend's selective recomputation further improves F1 under the same latency budget across most budgets on all 3 datasets, except at the tightest budgets below about 75\,ms on HotpotQA, while combining with KVLink yields mixed results.
At peak, \ours{}+CacheBlend reaches an F1 score of 55.8, 49.0, and 36.4 on HotpotQA, 2WikiMQA, and MuSiQue, compared with 51.4, 44.5, and 32.2 for \ours{} alone, and it also leads under the 100\,ms budget (Table~\ref{tab:latency}).
In contrast, \ours{}+KVLink changes peak F1 by only $+$0.9, $-$0.1, and $-$0.2 points (52.3, 44.4, and 32.0), and trails \ours{} alone under tight budgets on 2WikiMQA and MuSiQue.
Thus, integrating \ours{} with CacheBlend provides a new way to further improve accuracy under a fixed latency budget.

\section{Successful and Failed Inference Examples}
\label{app:inference-cases}

To illustrate how answer generation with nuggets in \ours{} works, we categorize the success cases from LongBench HotpotQA in Figure~\ref{fig:good-cases} (Type-1 to Type-3) and weak or failed cases in Figure~\ref{fig:bad-cases} (Type-1$'$ to Type-3$'$).
For those positive cases, our cache-stitching pipeline enables robust semantic inference and knowledge recovery within a highly compressed footprint.
However, these failure trajectories also illustrate that aggressive nugget slicing can sometimes amplify surface entity biases or omit contiguous linking paths during the retrieval and alignment stages, providing some insights for future optimizations.
Even with such failed cases, our evaluation in Section~\ref{sec:experimental_results} shows that \ours{} still delivers a strong F1-score on average for three test datasets and outperforms the baselines under a targeted latency budget.

\begin{figure*}[htbp]
\centering
\fbox{\parbox{0.98\textwidth}{\footnotesize\raggedright

\textbf{\large Type-1: Lexical match.}\quad
\begin{itemize}[leftmargin=1.5em, topsep=1pt, itemsep=0pt, label={}]
  \item \textit{Q: ``Which is a flowering plant, Pueraria or Pleiospilos?''} \hfill \textbf{Gold:} \textcolor{blue}{Pleiospilos} \quad \textbf{Pred:} \textcolor{blue}{pleiospilos} ($\checkmark$)
  \item \textbf{Top-5 chunks:} (1)~[0.593]~\texttt{chunk\_260} \textit{Pleiospilos} --- \textit{Pleiospilos is a genus of succulent flowering plants of the family Aizoaceae\ldots}\ \ (2)~[0.544]~\texttt{chunk\_255} \textit{Pueraria} --- \textit{Pueraria is a genus of 15--20 species of legumes\ldots}\ \ (3)~[0.535]~\texttt{chunk\_259} \textit{Pleiospilos simulans}\ \ (4)~[0.514]~\texttt{chunk\_261} \textit{Pleiospilos bolusii}\ \ (5)~[0.510]~\texttt{chunk\_258} \textit{Pleiospilos nelii}
  \item \textbf{Top-6 nuggets:} \textcolor{blue}{[0.635] ``Pleiospilos is a genus of succulent flowering plants of the family Aizoaceae, native to South Africa.''} \,|\, [0.591] ``Pleiospilos compactus'' \,|\, [0.574] ``Pleiospilos bolusii, the mimicry plant, is a species of flowering plant in the family Aizoaceae'' \,|\, [0.567] ``Pleiospilos nelii'' \,|\, [0.560] ``Pueraria is a genus of 15--20 species of legumes native to Asia.'' \,|\, [0.554] ``Pleiospilos bolusii''
\end{itemize}

\medskip\hrule\medskip

\textbf{\large Type-2: Semantic-only nugget.}\quad
\begin{itemize}[leftmargin=1.5em, topsep=1pt, itemsep=0pt, label={}]
  \item \textit{Q: ``Robbie Tucker plays in what series that follows a group of friends who run an Irish bar?''} \hfill \textbf{Gold:} \textcolor{blue}{It's Always Sunny in Philadelphia} \quad \textbf{Pred:} \textcolor{blue}{it's always sunny in philadelphia} ($\checkmark$)
  \item \textbf{Top-5 chunks:} (1)~[0.501]~\texttt{chunk\_780} \textit{IASIP / Paddy's Pub}\ \ (2)~[0.487]~\texttt{chunk\_777} \textit{Robbie Tucker} --- \textit{\ldots Tucker has also starred on \ldots It's Always Sunny in Philadelphia\ldots}\ \ (3)~[0.480]~\texttt{chunk\_3753} \textit{distractor}\ \ (4)~[0.478]~\texttt{chunk\_778} \textit{IASIP overview}\ \ (5)~[0.476]~\texttt{chunk\_786} \textit{Paddy's Pub location}
  \item \textbf{Top-6 nuggets:} \textcolor{blue}{[0.640] ``The series follows the exploits of `The Gang', a group of narcissistic and sociopathic friends who run the Irish dive bar Paddy's Pub in South Philadelphia, Pennsylvania''} \,|\, [0.496] ``The Gang originally consists of janitor Charlie Kelly\ldots'' \,|\, [0.495] ``Robbie Tucker (born April 5, 2001)'' \,|\, [0.474] ``Glenn Howerton as Dennis Reynolds, co-owner and the main bartender of Paddy's Pub'' \,|\, [0.462] ``In season 2, they are joined by Frank Reynolds (Danny DeVito)\ldots'' \,|\, [0.443] ``When the bar is open, they shirk their respective jobs' responsibilities and choose to drink instead.''
  \item \textbf{Note:} No nugget text contains the literal string ``It's Always Sunny in Philadelphia'' --- model infers the title from the show's plot description (Paddy's Pub, the Gang, South Philadelphia).
\end{itemize}

\medskip\hrule\medskip

\textbf{\large Type-3: Retrieval fails, model recovers.}\quad
\begin{itemize}[leftmargin=1.5em, topsep=1pt, itemsep=0pt, label={}]
  \item \textit{Q: ``Which ocean would the ingredients likely come from for the She-crab soup?''} \hfill \textbf{Gold:} \textcolor{blue}{Atlantic Ocean} \quad \textbf{Pred:} \textcolor{blue}{atlantic ocean} ($\checkmark$)
  \item \textbf{Top-5 chunks:} (1)~[0.695]~\texttt{chunk\_5252} \textit{She-crab soup} --- \textit{\ldots made of milk or heavy cream, crab or fish stock, Atlantic blue crab meat\ldots}\ \ (2)~[0.496]~\texttt{chunk\_3648} \textit{Crab trap}\ \ (3)~[0.455]~\texttt{chunk\_5254} \textit{Watercress soup}\ \ (4)~[0.436]~\texttt{chunk\_5247} \textit{Spinach soup}\ \ (5)~[0.421]~\texttt{chunk\_5967} \textit{Mildred Brown food columnist}
  \item \textbf{Top-6 nuggets:} [0.699] ``the orange crab roe comprise a chief ingredient in traditional she-crab soup.'' \,|\, [0.691] ``soup is named for the `she-crab', or female crab\ldots'' \,|\, \textcolor{blue}{[0.664] ``She-crab soup is a rich soup, similar to bisque, made of milk or heavy cream, crab or fish stock, Atlantic blue crab meat, and (traditionally) crab roe\ldots''} \,|\, [0.539] ``The soup is a regional specialty from the South Carolina Lowcountry.'' \,|\, [0.518] ``The crab pot changed the way crabs are harvested on the Chesapeake Bay.'' \,|\, [0.493] ``Crab has been a viable food source since Native Americans lived and fished on the Delmarva Peninsula.''
  \item \textbf{Note:} The literal answer ``Atlantic Ocean'' appears in no retrieved chunk or nugget. Only ``Atlantic blue crab'' (a species name) and geographic cues (South Carolina Lowcountry, Chesapeake Bay, Delmarva Peninsula) appear. Model bridges species name + east-coast geography $\rightarrow$ Atlantic Ocean.
\end{itemize}
}}
\caption{Three success case types of \ours{} on HotpotQA. All retrieval scores and IDs are from an actual \embed{} run. Predictions are from the model trained with SFT (\texttt{original\_offset}, $k_c=5$, $k=6$). \textcolor{blue}{Blue} marks the gold answer or its semantic cue.}
\label{fig:good-cases}
\end{figure*}

\begin{figure*}[htbp]
\centering
\fbox{\parbox{0.98\textwidth}{\footnotesize\raggedright

\textbf{\large Type-1$'$: Misled despite full evidence.}\quad
\begin{itemize}[leftmargin=1.5em, topsep=1pt, itemsep=0pt, label={}]
  \item \textit{Q: ``Gary L. Bennett was a part of the space missions that have a primary destination of what celestial body?''} \hfill \textbf{Gold:} \textcolor{blue}{Sun} \quad \textbf{Pred:} \textcolor{red}{jupiter} ($\times$)
  \item \textbf{Top-5 chunks:} (1)~[0.607]~\texttt{chunk\_107} \textit{Gary L. Bennett bio}\ \ (2)~[0.494]~\texttt{chunk\_108} \textit{Bennett UN delegations}\ \ (3)~[0.470]~\texttt{chunk\_135} \textit{Ulysses (spacecraft)} --- \textit{\ldots primary mission was to orbit the Sun\ldots}\ \ (4)~[0.454]~\texttt{chunk\_128} \textit{distractor}\ \ (5)~[0.436]~\texttt{chunk\_139} \textit{Pioneer H / Out-Of-The-Ecliptic}
  \item \textbf{Top-6 nuggets:} \textcolor{blue}{[0.525] ``was a robotic space probe whose primary mission was to orbit the Sun and study it at all latitudes.''} \,|\, \textcolor{red}{[0.524] ``went to Jupiter, Saturn, Uranus, Neptune and beyond''} \,|\, \textcolor{red}{[0.494] ``and on the New Horizons mission to Jupiter.''} \,|\, \textcolor{red}{[0.484] ``it was propelled on a trajectory to Jupiter by a combination of solid rocket motors''} \,|\, [0.475] ``Voyager, Galileo, and Ulysses space missions'' \,|\, [0.472] ``KELT is a terrestrial telescope mission designed to search for transiting systems''
  \item \textbf{Note:} The gold answer ``Sun'' appears in the top-1 nugget, but four out of six nuggets mention ``Jupiter''. Model is misled by surface frequency despite the highest-scoring nugget pointing to the correct answer.
\end{itemize}

\medskip\hrule\medskip

\textbf{\large Type-2$'$: Plausible alternative entity.}\quad
\begin{itemize}[leftmargin=1.5em, topsep=1pt, itemsep=0pt, label={}]
  \item \textit{Q: ``Name a member of a British-American supergroup who recorded a version of Nobody's Child in 1990''} \hfill \textbf{Gold:} \textcolor{blue}{Bob Dylan} \quad \textbf{Pred:} \textcolor{red}{george harrison} ($\times$)
  \item \textbf{Top-5 chunks:} (1)~[0.529]~\texttt{chunk\_642} \textit{Wilburys pseudonyms}\ \ (2)~[0.519]~\texttt{chunk\_634} \textit{Nobody's Child album} --- \textit{compiled by \ldots George Harrison}\ \ (3)~[0.485]~\texttt{chunk\_638} \textit{Traveling Wilburys} --- \textit{Bob Dylan, George Harrison, Jeff Lynne, Roy Orbison and Tom Petty}\ \ (4)~[0.481]~\texttt{chunk\_643}\ \ (5)~[0.471]~\texttt{chunk\_637} \textit{Nobody's Child song disambiguation}
  \item \textbf{Top-6 nuggets:} [0.565] ``Nobody's Child: Romanian Angel Appeal, a 1990 charity album'' \,|\, \textcolor{red}{[0.549] ``his presence on Nobody's Child was reflected in the recordings submitted by Simon, Clapton and Eddy''} \,|\, \textcolor{red}{[0.534] ``a cover of `Nobody's Child', which the band recorded for Olivia Harrison's Romanian Angel Appeal charity project.''} \,|\, [0.533] ``Nobody's Child (Penny McLean song), 1976'' \,|\, [0.524] ``covered by Tony Sheridan and the Beatles (1964), the Traveling Wilburys (1990), and others'' \,|\, [0.516] ``Nobody's Child, a song by Electric Light Orchestra from Eldorado, 1974''
  \item \textbf{Note:} The Wilburys roster (``Bob Dylan, George Harrison, Jeff Lynne, Roy Orbison and Tom Petty'') exists in chunk\_638 but no nugget extracted it. Only Harrison-centric nuggets survive (Olivia Harrison, Harrison as compiler). Model picks Harrison --- valid Wilburys member, but mismatches the canonical reference.
\end{itemize}

\medskip\hrule\medskip

\textbf{\large Type-3$'$: Retrieval fails, model fabricates plausible.}\quad
\begin{itemize}[leftmargin=1.5em, topsep=1pt, itemsep=0pt, label={}]
  \item \textit{Q: ``Prior to playing for Michigan State, Keith Nichol played football for a school located in what city?''} \hfill \textbf{Gold:} \textcolor{blue}{Norman} \quad \textbf{Pred:} \textcolor{red}{lowell} ($\times$)
  \item \textbf{Top-5 chunks:} (1)~[0.610]~\texttt{chunk\_91} \textit{Keith Nichol} --- \textit{played college football for the University of Oklahoma\ldots attended Lowell High School in Lowell, Michigan}\ \ (2)~[0.498]~\texttt{chunk\_92} \textit{Nichol sophomore season}\ \ (3)~[0.478]~\texttt{chunk\_67} \textit{John Macklin}\ \ (4)~[0.476]~\texttt{chunk\_93} \textit{Nichol senior season}\ \ (5)~[0.473]~\texttt{chunk\_105} \textit{1966 Notre Dame vs.\ MSU}
  \item \textbf{Top-6 nuggets:} \textcolor{red}{[0.567] ``played college football for the University of Oklahoma and Michigan State University''} \,|\, [0.553] ``transferred to Michigan State'' \,|\, [0.519] ``Both schools shared the MacArthur Bowl'' \,|\, [0.514] ``Nichol played an important role as receiver for the Spartans during the 2011 season'' \,|\, \textcolor{red}{[0.514] ``attended Lowell High School in Lowell, Michigan''} \,|\, [0.511] ``coached football at a boys' school at Pawling, New York.''
  \item \textbf{Note:} Gold ``Norman'' (Oklahoma's city) appears in no chunk or nugget --- retrieval brings ``University of Oklahoma'' but no separate passage about Norman, OK. Model picks ``Lowell'' (from the Lowell HS nugget), interpreting ``school'' as high school instead of the intended college reading.
\end{itemize}
}}
\caption{Three failed case types of \ours{} on HotpotQA, mirroring the success cases in Figure~\ref{fig:good-cases}. \textcolor{red}{Red} marks the misleading nugget/answer; \textcolor{blue}{blue} marks the gold answer (when present).}
\label{fig:bad-cases}
\end{figure*}

\end{document}